\documentclass{article}

\usepackage[final,nonatbib]{nips_generic}

\usepackage[utf8]{inputenc} 
\usepackage[T1]{fontenc}    
\usepackage{hyperref}       
\usepackage{url}            
\usepackage{booktabs}       
\usepackage{amsfonts}       
\usepackage{nicefrac}       
\usepackage{microtype}      
\usepackage{algpseudocode}
\usepackage{algorithm} 
\usepackage{amsmath}
\usepackage{bbm}
\usepackage{dsfont}
\usepackage{stmaryrd}
\usepackage{color}
\usepackage{graphicx,graphics,subcaption}
\usepackage[normalem]{ulem}

\DeclareUnicodeCharacter{00A0}{~}

\newcommand{\revOneAdd}[1]{#1}
\newcommand{\revOneDel}[1]{}

\newcommand{\revTwoAdd}[1]{#1}
\newcommand{\revTwoDel}[1]{}

\newmuskip\pFqmuskip

\newcommand*\pFq[6][8]{%
  \begingroup 
  \pFqmuskip=#1mu\relax
  \mathcode`\,=\string"8000
  \begingroup\lccode`\~=`\,
  \lowercase{\endgroup\let~}\pFqcomma
  {}_{#2}F_{#3}{\left[\genfrac..{0pt}{}{#4}{#5};#6\right]}%
  \endgroup
}
\newcommand{\pFqcomma}{\mskip\pFqmuskip}

\title{Taming Non-stationary Bandits:\\ A Bayesian Approach}

\author{
    Vishnu Raj \qquad Sheetal Kalyani \\
    Department of Electrical Engineering \\
    Indian Institute of Technology, Madras \\
    \texttt{\{ee14d213,skalyani\}@ee.iitm.ac.in}
}

\begin{document}
    \maketitle

    \begin{abstract}
        We consider the multi armed bandit problem in non-stationary environments. Based on the
        Bayesian method, we propose a variant of Thompson Sampling which can be used in both
        rested and restless bandit scenarios. Applying discounting to the parameters of prior
        distribution, we describe a way to systematically reduce the effect of past observations.
        Further, we derive the exact expression for the probability
        of picking sub-optimal arms. By increasing the exploitative value of Bayes' samples, we
        also provide an optimistic version of the algorithm.
        Extensive empirical analysis is conducted under various scenarios to validate the utility
        of proposed algorithms. A comparison study with various state-of-the-arm algorithms is also
        included.
    \end{abstract}

\section{Introduction}
    \textbf{Background and Motivation.}
    Multi armed bandit (MAB) is a well known paradigm for sequential decision making under uncertainty
    and partial feedback. In such scenarios, there exists a tension between \textit{exploring} different options
    at the cost of losing the optimal action and \textit{exploiting} the information already acquired at the
    cost of ignoring the uncertainty. To minimize the loss that can possibly be incurred 
    by choosing suboptimal options, the decision maker has to find a balance between the exploration and
    exploitation phases. This problem of optimally allocating information acquisition efforts to exploration
    and exploitation phases is originally proposed in \cite{Thompson1933} in the context of clinical 
    trials. The formalization of the problem is done in \cite{Herbert1952}, in which each action is viewed
    as an arm indexed by $i = 1,\ldots,K$, with an unknown probability distribution $\nu_i$, and the pay-off 
    from arm $i$ at each instant $t$,
    $X_{i,t}$, as an independent draw from $\nu_i$. Depending on the assumed nature of reward structure, MAB 
    problems can be divided into stationary bandits and non-stationary bandits. 
    
    Stationary bandit formulation assumes the underlying pay-off probability  distribution of each option 
    to be stationary. Hence, statistical properties such as mean, variance etc., remain constant for the entire 
    period of interest. Most widely used metric to compare the desirability of an option over another is its 
    mean pay-off. Because of the stationary behavior of pay-off, we can hope to converge to 
    the best option, atleast asymptotically. Seminal work in  \cite{Lai1985} introduced the technique of
    upper confidence bounds for the asymptotic analysis of loss incurred while playing a suboptimal option.
    Notable works in this area include Gittins indices \cite{Gittins1979a,Gittins1979b,Weber1992},
    upper confidence bound based policies \cite{Auer2002,Auer2010}, probability matching techniques 
    \cite{Chapelle2011,Agrawal2012b}. Recent additions to this family are POKER \cite{Vermorel2005}, KL-UCB algorithm \cite{Garivier2011}, 
    Bayes-UCB \cite{Kaufmann2012} etc., which provide tight performance bounds.
    
    There exists a different kind of bandit formulation where the reward generating process is no longer
    assumed to be stationary. This makes the
    problem harder than stationary bandits as there is no single optimal option to which an algorithm can converge. 
    Adversarial bandit formulation is one of the strongest generalizations of this case, where the reward
    generating process is controlled by an adversary for worst-case per play \cite{Bubeck2012}. Although it may
    seem hopeless to play such a game, randomizing the process of decision making is proposed as
    a method for minimizing regret \cite{Auer1995,Auer2002b}. Efforts have also gone
    into unifying these two separate worlds of stochastic and adversarial bandits \cite{Bubeck2012a,Seldin2014}.
    Another formulation, commonly referred to as \textit{Non-Stationary} bandits, does not assume a 
    stationary reward generating process. Dealing with reward processes that can evolve over time, this line
    of work is of great importance in modelling real world processes and is currently a very active research
    area \cite{Hartland2006,Slivkins2008a,Garivier2011b,Besbes2014a,Neu2015a}. For more details about multi armed
    bandits, interested readers are referred to \cite{Burtini2015}.
    
    Thompson Sampling is one of the oldest algorithms proposed for the trade-off between exploration and 
    exploitation \cite{Thompson1933}. It works by selecting an arm to pull according to its
    probability of being the optimal. 
    Recent studies showing strong empirical results \cite{Chapelle2011,Scott2010} followed by solid theoretical 
    guarantees \cite{Agrawal2012b,Kaufmann2012,Agrawal2013} have rekindled the interest in Thompson Sampling method.
    Another line of work has come up with information theoretic analysis of Thompson Sampling under 
    stationary environments \cite{Russo2015,Zhou2015}. In \cite{Leike2016}, authors have proved
    that a variant of Thompson Sampling is asymptotically optimal in non-parametric reinforcement learning
    under countable classes of general stochastic environments. However, there exists very few results 
    for the analysis of Thompson Sampling in non-stationary cases.

    \textbf{Related Works.} One of the earliest works in dynamic bandits with abrupt changes in the reward generation 
    process is the algorithm 
    \textit{Adapt-EvE} proposed in \cite{Hartland2006}. It uses a change point detection technique to detect any 
    abrupt change in the environment and utilizes a meta bandit formulation for \textit{exploration-exploitation dilemma} 
    once change is detected. Authors of \cite{Slivkins2008a,Slivkins2007} considered a dynamic bandit setting where the reward
    evolves as Brownian motion and provided results of regret linear in time horizon $T$. \revOneDel{A method named \textit{Dynamic Thompson
    Sampling} is introduced in \cite{Gupta2011a} and is empirically shown to be performing better in an environment where reward
    probability variations are Brownian in nature. }
    In an effort to combine both stochastic and adversarial regimes, authors of \cite{Seldin2014} proposed
    EXP3++ algorithm which achieves almost optimal performance in both cases. A popular belief was that in non-stationary bandits,
    for high-confidence performance guarantees, the player has to sample all the arms uniformly atleast $\Omega(\sqrt{T})$
    times. An undesirable effect of this is the growth of regret at $O(\sqrt{T})$. But, \cite{Neu2015a} showed 
    theoretically that this need not be the case and strong guarantees can be derived with high probability without
    this requirement. Based
    on this observation, authors proposed a variant of \textit{EXP3} algorithm with \textit{Implicit Exploration}, called \textit{EXP3-IX}.
    Interestingly, however, empirical studies showed that \textit{EXP3-IX} also sampled
    arms roughly $\sqrt{T}$ number of times.
    \revOneDel{In \cite{Xu2013}, authors discusses about Thompson Sampling technique
    for dynamic contextual bandits and discusses about the impact of discounted decaying of past samples in dynamic
    systems.}
    
    \revOneAdd{In most of the previously discussed algorithms, the ability to respond to the changing environment is made
    possible either by resetting the algorithm at suitable points or by allowing explicit exploration. An
    alternate way to tackle the problem of non-stationarity is to reduce the impact of past observations in the
    current prediction in a systematic manner. By discounting the effect of past observations suitably, the predictions
    from the model can be made based on more recent samples. In the context of dynamic bandits, this concept of applying
    an exponential filtering to past observations is suggested in \cite{Garivier2011b}. Extending the idea to
    Bayesian methods, \cite{Gupta2011a} proposed  Dynamic Thompson Sampling (Dynamic TS). 
    By assuming a Bernoulli
    bandit environment where the success probability evolves as a Brownian motion, authors suggest to decay the effect
    of past observations in the posterior distribution of the arm being updated. This is
    done by applying an exponential filtering to the past observations. However, by only discounting past
    observations of the pulled arm, this algorithm is more suitable for a \textit{rested bandit} case where the underlying distribution
    changes only when the arm is played. But in the case of \textit{restless bandits}, where the underlying distribution
    of all the arms changes at every time instant, Dynamic TS may perform poorly. One of the trivial
    cases where this can happen is when the past optimal arm remains stationary during the game, and a previously
    suboptimal non-stationary arm becomes optimal. Dynamic TS will find it difficult to switch to the new optimal arm, 
    as the statistical
    properties of past optimal arm remains unchanged, thus missing a chance to explore any other suboptimal arm.}
    
    
    \revOneAdd{One of the fundamental questions that need to be answered while moving away from the well established
    stationary bandits is how to ascertain the performance of candidate policies. The absence of a single
    optimal arm for the entire game makes it difficult to set a proper benchmark for the performance of proposed
    solutions. In the adversarial bandit setting, the benchmark to be compared against is taken as the best arm at 
    hindsight, which represents a \textit{static oracle}. Another benchmark that can be used is a \textit{dynamic
    oracle} who selects the optimal option at every instant. However, the use of \textit{dynamic oracle} as a benchmark
    is less discussed in literature because of the difficulty in mathematical analysis. As mentioned in 
    \cite{Besbes2014a}, \textit{static oracle} can perform quite poorly when compared to \textit{dynamic oracle}.
    However, by assuming subtle structures in the variations of reward generating process, \cite{Besbes2014a} was
    successful in establishing bounds on minimal achievable regret against a \textit{dynamic oracle} and developed
    a near optimal policy, REXP3. However, in real applications, it is difficult to know about structures in 
    the variations of environment.}

    \textbf{Main Contributions.} \revOneDel{This paper proposes a variant of Thompson sampling for non-stationary 
    environments. The proposed 
    method can recover from being stuck with a previously optimal arm at the cost of more exploration. We
    provide simulation results in a wide range of non-stationary environments and empirically show that the
    proposed method can perform better than conventional Thompson Sampling.}
    \revOneAdd{This paper proposes a Bayesian bandit algorithm for non-stationary environments. Derived from
    the popular Thompson Sampling (TS) algorithm, our proposed method - \emph{Discounted Thompson Sampling (dTS)} - works
    by discounting the effect of past observations. 
    Even though a similar technique is proposed in Dynamic TS \cite{Gupta2011a}, our method differs differs from
    it in two aspects. First, we update parameters of all posterior distributions at every timestep, while Dynamic
    TS updates the parameters only for the arm it played. Next, our algorithms applies
    the filtering at all timesteps, but DTS applies it to arm only after the condition $\alpha_k+\beta+k > C$ is
    met for the played arm. This exponential filtering technique increases
    the variance of the prior distribution maintained for all the arms, while keeping the mean almost constant unless that
    arm is played.
    By increasing the variance of all arms, we increase the probability of picking past 
    inferior arms for exploration. However, by keeping the mean almost constant, we restrain the algorithm from 
    picking inferior arms too often. This makes the algorithm suitable for many non-stationary cases, including the 
    notoriously difficult restless bandit formulation. Further, inspired from the optimistic Bayesian approaches 
    in bandit problems, we add an optimistic version of dTS, Discounted Optimistic Thompson Sampling (dOTS). 
    Numerical verification of the performance of the proposed algorithms is conducted and a comparison with various 
    state-of-the-art algorithms is provided for a variety of worst-case scenarios with \textit{dynamic oracle}.}
    
\section{Problem Formulation}
    Let $\mathcal{K} = \{1,\ldots,K\}$ denotes the set of arms available to the decision maker. Let the
    horizon of the game be denoted by $T$. Hence, at every time instant $t \in \{1,\ldots,T\}$, the decision maker
    has to choose one arm $k \in \mathcal{K}$ to play. Let $X_{k,t}$ denote the reward obtained by pulling the
    $k^{th}$ arm at $t^{th}$ time instant. This instantaneous reward $X_{k,t}$ is a Bernoulli random variable with mean,
    $\mu_{k,t} = \mathbb{E}X_{k,t}$. Best arm at any instant $t$ is the arm with highest expected reward at that
    instant and is denote by $\mu^{*}_{t} = \underset{k\in\mathcal{K}}{\max}\{\mu_{k,t}\}$.
    
    In non-stationary systems, the expected rewards will evolve over time. This evolution of reward probabilities
    can either be abrupt or can show a trend. This sequence of rewards from $k^{th}$ arm 
    is denoted by $\mu^{k} = 
    \left\{\mu_{k,t}\right\}_{t=1}^{T}$. Let $\mu = \left\{ \mu^{k} \right\}_{k\in\mathcal{K}}$ denote 
    the vector of sequence of rewards of all arms.
    
    \revOneAdd{Let $\mathcal{P}$ denote the family of admissible policies and let $\pi \in \mathcal{P}$ denote a 
    candidate policy that selects an arm to pull during the game.} At each instant, policy
    $\pi$ selects an arm $I^{\pi}_{t}$ based on the initial prior $U$ and the past observations 
    $\{X_{I^\pi_{n},n}\}_{n=1}^{t-1}$. \revOneDel{Also $\pi \in \mathcal{P}$ where $\mathcal{P}$ is the family of admissible
    policies.} If we denote $\pi_t$ to be the policy at instant $t$, then
    \begin{align}
        \pi_t &= \begin{cases}
                \pi_1(U) & ; t = 1 \\
                \pi_t(U,X_{I^\pi_1,1},\ldots,X_{I^\pi_{t-1},t-1}) & ; t \geq 2
            \end{cases}
    \end{align}
    
    Thompson Sampling proceeds by maintaining a prior distribution over the success probability of each Bernoulli 
    arm and sampling
    from this prior distribution for selecting the arm to play. Beta distribution is selected as the prior distribution
    because it is the conjugate prior of Bernoulli distribution. The Beta distribution has two parameters, $\alpha$ and $\beta$, which gets
    updated according to the rewards seen during the game.
    For both dTS and dOTS, we have $\alpha, \beta \in 
    \mathbb{R}_{>0}^{|\mathcal{K}|}$ along with the discounting factor $\gamma \in [0,1]$ and hence 
    $\pi_T : [0,1] \times \mathbb{R}_{>0}^{|\mathcal{K}|} \times 
    \mathbb{R}_{>0}^{|\mathcal{K}|} \times \{0,1\}^{T-1} \rightarrow \mathcal{K} $.
    
    \revOneAdd{As mentioned earlier, one of the fundamental questions that arises is the optimal policy 
    against which the candidate policies can be benchmarked. 
    We follow the notion of dynamic oracle \cite{Besbes2014a} as the optimal policy for comparing
    the performance of the algorithms. Dynamic oracle optimizes the expected reward
    at each time instant over all possible actions.} \textit{Regret} $\mathcal{R}^{\pi}(T)$,
    is defined as the difference between the
    expected cumulative reward from dynamic oracle and the expected reward from the policy under test, $\pi$.
    Hence, regret is defined as
    \begin{align}
        \mathcal{R}^{\pi}(T) &= \sum \limits_{t=1}^{T} \mu_t^{*} 
                - \mathbb{E}_{\pi,\mu} \left[ \sum \limits_{t=1}^{T} X_{I^{\pi}_{t},t} \right]
    \end{align}
    where the expectation $\mathbb{E}_{\pi,\mu}[\cdot]$ is taken over both the randomization in the policy and
    the randomization in the environment. To provide a normalized version, our experiments use $\frac{1}{T} \mathcal{R}^{\pi}
    (T)$ as the performance metric.

\section{Discounted Thompson Sampling}
    \revOneAdd{This section introduces the \textit{Discounted Thompson Sampling (dTS)} algorithm. As mentioned
    earlier, the key idea used in designing the algorithm is to systematically increase the variance of
    the prior distributions maintained for unexplored
    arms. Hence, the probability of picking them will get increased. Another
    key feature is that, while modifying the distribution to increase its variance, the mean of the distributions
    are kept almost constant between the plays. 
    Based on the Bayesian approach, the
    algorithm will sample from this modified distribution to select the arm to play.
    The mean will get modified only for the arm which is played. The 
    algorithm is listed in Algorithm \ref{alg:dTS}.}
    \begin{algorithm}[H]
        \caption{Discounted Thompson Sampling (dTS)}    \label{alg:dTS}
        \begin{algorithmic}[0]
            \State \textbf{Parameters :} $\gamma \in (0,1]$, 
                        $\alpha_0,\beta_0 \in \mathbb{R}_{\geq0}$,
                        $K = |\mathcal{K}| \geq 2$
            \State \textbf{Initialization :} $S_k = 0, \quad F_k = 0 \qquad \forall k \in \{1,\ldots,K\}$
            \For{ t = 1, 2, \ldots, T}
                \For{ $k = 1, \ldots, K$}
                    \State $\theta_k(t) \sim Beta(S_k+\alpha_0,F_k+\beta_0)$
                \EndFor
                \State Play arm $I^{\pi}_{t}:= \arg \max _k\:\theta_k(t)$ and observe reward $\tilde{r}_t$.
                \State Perform a Bernoulli trial with success probability $\tilde{r}_t$ and observe output $r_t$.
                \State Update $S_{I^{\pi}_{t}} \gets \gamma S_{I^{\pi}_{t}} + r_t$ and $F_{I^{\pi}_{t}} \gets \gamma F_{I^{\pi}_{t}} + (1-r_t)$ .
                \State Update $S_k \gets \gamma S_k$ and $F_k \gets \gamma F_k;$ $\forall\: k \neq I^{\pi}_{t}$.
            \EndFor
        \end{algorithmic}
    \end{algorithm}
    \revOneAdd{Here, $\alpha_0$ and $\beta_0$ are used as the initial values of parameters for the Beta prior
    distributions. Comparing to TS, dTS discounts the past value of $S_k$ and $F_k$ before updating
    it with the current reward. This discounting is performed even for the arms which are not played at this time 
    instant.} By setting $\alpha_0 = 1$ and $\beta_0 = 1$ along with $\gamma = 1$, 
    we get the traditional Thompson Sampling.
    
    Let $\llbracket \mathcal{A} \rrbracket$ denote the indicator function that event $\mathcal{A}$ has occurred. From 
    Algorithm \ref{alg:dTS}, we can get the update equation for parameters of prior distributions as
        $ S_{k,t+1} \gets \gamma \cdot S_{k,t} + \llbracket I^{\pi}_{t}=k \rrbracket \llbracket r_t=1 \rrbracket $ and
        $ F_{k,t+1} \gets \gamma \cdot F_{k,t} + \llbracket I^{\pi}_{t}=k \rrbracket \llbracket r_t=0 \rrbracket $.
    \revOneAdd{By taking expectation over the randomization in the algorithm and the environment, we can write the expected
    values of prior parameters as}
    \begin{align}
        \mathbb{E}S_{k,t+1} = \gamma \mathbb{E}S_{k,t} + \mu_{k,t} \mathbb{P}(I^\pi_t = k) \qquad \text{and} \qquad
        \mathbb{E}F_{k,t+1} = \gamma \mathbb{E}F_{k,t} + (1-\mu_{k,t}) \mathbb{P}(I^\pi_t = k),
    \end{align}
    \revOneAdd{where $\mathbb{P}(I^\pi_t = k)$ is the probability of selecting arm $k$ at instant $t$ given all the
    history $\{r_{I^\pi_{t},n}\}_{n=1}^{t-1}$.}
    
    \revOneAdd{Neglecting the effect of $\alpha_0$ and $\beta_0$, for the arms which are not played at this instant,
    we have the posterior mean as}
    \begin{align}
        \mu_{k,t+1} = \frac{S_{k,t+1}}{S_{k,t+1}+F_{k,t+1}} 
                    = \frac{ \gamma \cdot S_{k,t}}{\gamma \cdot S_{k,t}+\gamma \cdot F_{k,t}}
                    = \mu_{k,t}    \label{exp:dTS_mean}
    \end{align}
    and posterior variance as
    \begin{align}
        \sigma^2_{k,t+1} &= \frac{S_{k,t+1} \cdot F_{k,t+1}}{(S_{k,t+1} + F_{k,t+1})^2 (S_{k,t+1} + F_{k,t+1} + 1)}
                         = \frac{S_{k,t} \cdot F_{k,t}}{(S_{k,t} + F_{k,t})^2 (\gamma \cdot S_{k,t} + \gamma \cdot F_{k,t} + 1)} \nonumber \\
                         &= \frac{\mu_{k,t}(1-\mu_{k,t})}{\gamma \cdot S_{k,t} + \gamma \cdot F_{k,t} + 1}
                         \geq \frac{\mu_{k,t}(1-\mu_{k,t})}{ S_{k,t} + F_{k,t} + 1}
                         = \sigma^2_{k,t}. \label{exp:dTS_var}
    \end{align}
        
    \revOneAdd{From (\ref{exp:dTS_mean}) and (\ref{exp:dTS_var}), we can see that by discounting the past values of $S_{k}$ 
    and $F_{k}$ before updating the parameter of the posterior distribution, dTS algorithm is 
    able to increase the variance of the prior distributions while keeping the mean almost constant  
    for the arms which are not pulled.}
    
    \subsection{Being Optimistic}
    Inspired from the optimistic Bayesian sampling introduced in \cite{May2012}, dTS can be modified to
    incorporate optimism while sampling from the prior distribution for picking the arm to play. In this optimistic
    approach, \textit{Discounted Optimistic Thompson Sampling} (dOTS), 
    the samples from each prior distribution $\nu_k$ will get modified as
    \begin{align}
        \tilde{\theta}_k(t) &\sim Beta(S_k+\alpha_0,F_k+\beta_0) \nonumber \\
        \theta_k(t) &= \max\{ \mathbb{E}[\tilde{\theta}_k(t);\alpha_0,\beta_0,S_k,F_k], \tilde{\theta}_k(t) \}.
    \end{align}
    In dOTS, the prior samples are forced to have atleast its expected value, thus increasing the arms'
    exploitative value. Although this approach may seem counter-productive in non-stationary environments,
    empirical studies, provided later in this paper, show that dOTS can outperform dTS in non-stationary 
    environments. 

\section{Performance Analysis}
    \revOneAdd{Even though tight bounds on the performance of Thompson Sampling are available, all of them are derived
    based on the assumption of stationary environments and integer parameters \cite{Agrawal2012b,Kaufmann2012,Agrawal2013}. 
    In a stationary environment,
    we can divide the arms into two sets - one set of arms which have converged to the neighborhood of its
    actual mean and other set of arms which have not converged. By analyzing the probability with which arms
    from the second set will be picked, the regret of algorithm can be derived. However, in the non-stationary case,
    such a convergence is not meaningful. Further, by assuming the prior parameters to be integers, previous
    analysis of TS maps the probabilities to be Binomial distribution and hence, use the
    results from concentration inequalities to approximate the bounds. In the dTS and dOTS, the parameters
    need not be integers due to the discounting, which further complicates the analysis. Because of all these
    reasons, in the current work, we analyze the probability of picking a sub-optimal arm for the simple case
    of a two armed bandit.}
    \revOneAdd{In this section, 
    we provide the exact expression for the probability of picking a sub-optimal
    arm. To the best of our knowledge, this is the first time an exact expression for this is derived for
    non-integer parameters.}    
    We consider the simple case of
    two armed dynamic bandit, $\mathcal{K} = \{1,2\}$, in this section.

    \subsection{Probability of picking sub-optimal arm}
        Consider a two armed bandit with non-stationary reward structure. Without loss of generality, let
        arm with index $1$ be the optimal arm. Thompson sampling selects the next arm to play by drawing
        two independent samples from the Beta distributions maintained for each arm. Each arm will start
        with initial values of $\alpha_{i,0}$ and $\beta_{i,0}$, $ i = \{1,2\}$. Also denote the parameter
        values at time instant $t$ by $\alpha_{i,t}$ and $\beta_{i,t}$. We are interested in finding the 
        probability with which Thompson sampling will pick the suboptimal arm. Let $\theta_{1,t} \sim 
        Beta(\alpha_{1,t},\beta_{1,t})$ be the sample for optimal arm and $\theta_{2,t} \sim Beta(\alpha_{2,t},
        \beta_{2,t})$ for suboptimal arm. We are interested in finding $\mathbb{P}(\theta_{2,t} > \theta_{1,t}|
        \alpha_{1,t},\beta_{1,t},\alpha_{2,t},\beta_{2,t})$. Here on, we are dropping the sub-index $t$
        for notational simplicity.
        \begin{align}
            \mathbb{P}\left( \theta_{2} > \theta_{1}\right)
                &= \mathbb{P}\left( \frac{\theta_{1}}{\theta_{2}} < 1\right)    
        \end{align}
    
        From \cite{Pham-Gia2000}, we have
        \begin{align}
            \mathbb{P}\left( \frac{\theta_{1}}{\theta_{2}} = \omega \right) 
                &= \frac{B(\alpha_{1}+\alpha_{2},\beta_{2})}{B(\alpha_{1},\beta_{1})B(\alpha_{2},\beta_{2})}
                    \omega^{\alpha_{1}-1}                
                    \textstyle\pFq{2}{1}{\alpha_{1}+\alpha_{2},1-\beta_{1}}{\alpha_{1}+\alpha_{2}+\beta_{2}}{\omega}                
                    \qquad ;0 < \omega \leq 1
        \end{align}
        where $B(\alpha,\beta)$ is the Beta function and $\textstyle\pFq{2}{1}{}{}{}$ is Gauss hypergeometric function.
        
        The probability of picking a sub-optimal arm can now be written as
        \begin{align}
            \mathbb{P}\left(\theta_2 > \theta_1 \right)
                &= \mathbb{P}\left( \frac{\theta_1}{\theta_2} < 1 \right) \nonumber 
                 = \int_{0}^{1} \mathbb{P}\left( \frac{\theta_1}{\theta_2} = \omega \right) d\omega \nonumber \\
                &= \frac{B(\alpha_{1}+\alpha_{2},\beta_{2})}{B(\alpha_{1},\beta_{1})B(\alpha_{2},\beta_{2})}
                   \int_{0}^{1} \omega^{\alpha_{1}-1}                
                   \textstyle\pFq{2}{1}{\alpha_{1}+\alpha_{2},1-\beta_{1}}{\alpha_{1}+\alpha_{2}+\beta_{2}}{\omega} d\omega \nonumber \\
                &= \frac{B(\alpha_{1}+\alpha_{2},\beta_{2})}{B(\alpha_{1},\beta_{1})B(\alpha_{2},\beta_{2})}
                   \frac{1}{\alpha_1}
                   \textstyle\pFq{3}{2}{\alpha_{1},\alpha_{1}+\alpha_{2},1-\beta_{1}}{1+\alpha_{1},\alpha_{1}+\alpha_{2}+\beta_{2}}{1}
        \end{align}
        where $\textstyle\pFq{p}{q}{}{}{}$ is the \textit{Generalized Hypergeometric function}.
        
        These expressions hold for $\beta_0 > \frac{1}{2}$ (See supporting material for the details).

\section{Numerical Analysis}
    \revTwoDel{For the purpose of validating the proposed algorithm, this section presents a set of simulation studies. We have
    chosen a set of synthetic environments to closely represent worse case real world scenarios and the results
    of proposed algorithm compared against conventional Thompson Sampling\cite{Thompson1933} and REXP3 
    Algorithm\cite{Besbes2014a}. To test different variants of proposed
    method, we simulated Discounted Thompson Sampling(DTS) for two prior settings - $(a)\:\alpha_0 = 0, \beta_0 = 0,
    \bar{\alpha} = 1, \bar{\beta} = 1$ and $(b)\:\alpha_0 = 1, \beta_0 = 1, \bar{\alpha} = 0, \bar{\beta} = 0$. 
    Discounting factor, $\gamma$, for the experiments are selected according to environment. For REXP3 algorithm,
    the resetting period, $\Delta$, is set to exact period of swap of arms. }
    
    \revTwoAdd{This section includes results of numerical evaluation of the proposed algorithm in various synthetic environments.
    Results also include comparison of the proposed algorithm against various state-of-the-art algorithms. All the results
    provided are averaged over 5000 independent runs. }
    
    \subsection{Regret over time}
\begin{figure}[h]
            \begin{subfigure}[h]{0.33\textwidth}
                \includegraphics[width=\linewidth]{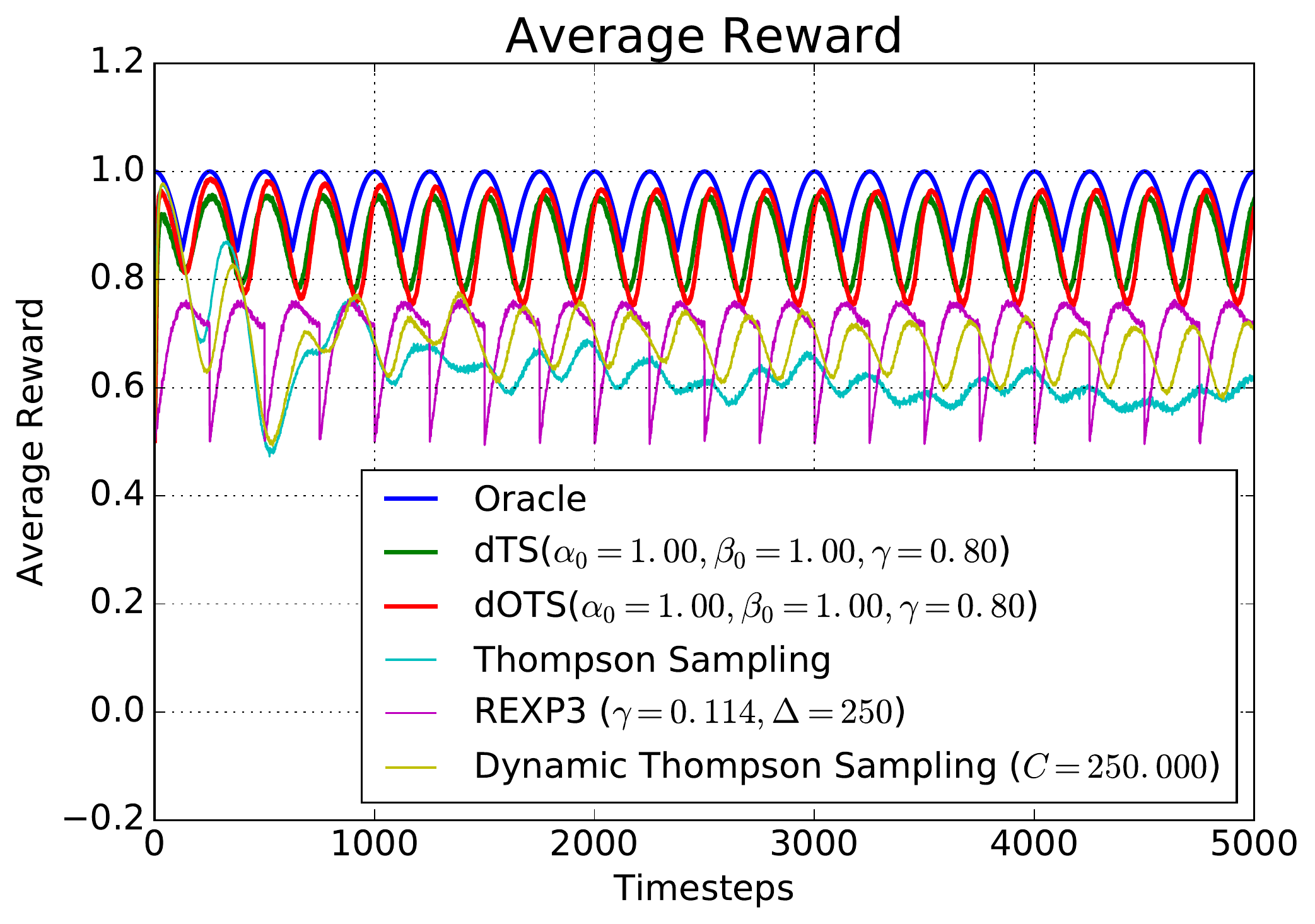}
                \caption{Slow Varying Environment}
                \label{fig:sve_rewards}
            \end{subfigure}
            \begin{subfigure}[h]{0.33\textwidth}
                \includegraphics[width=\linewidth]{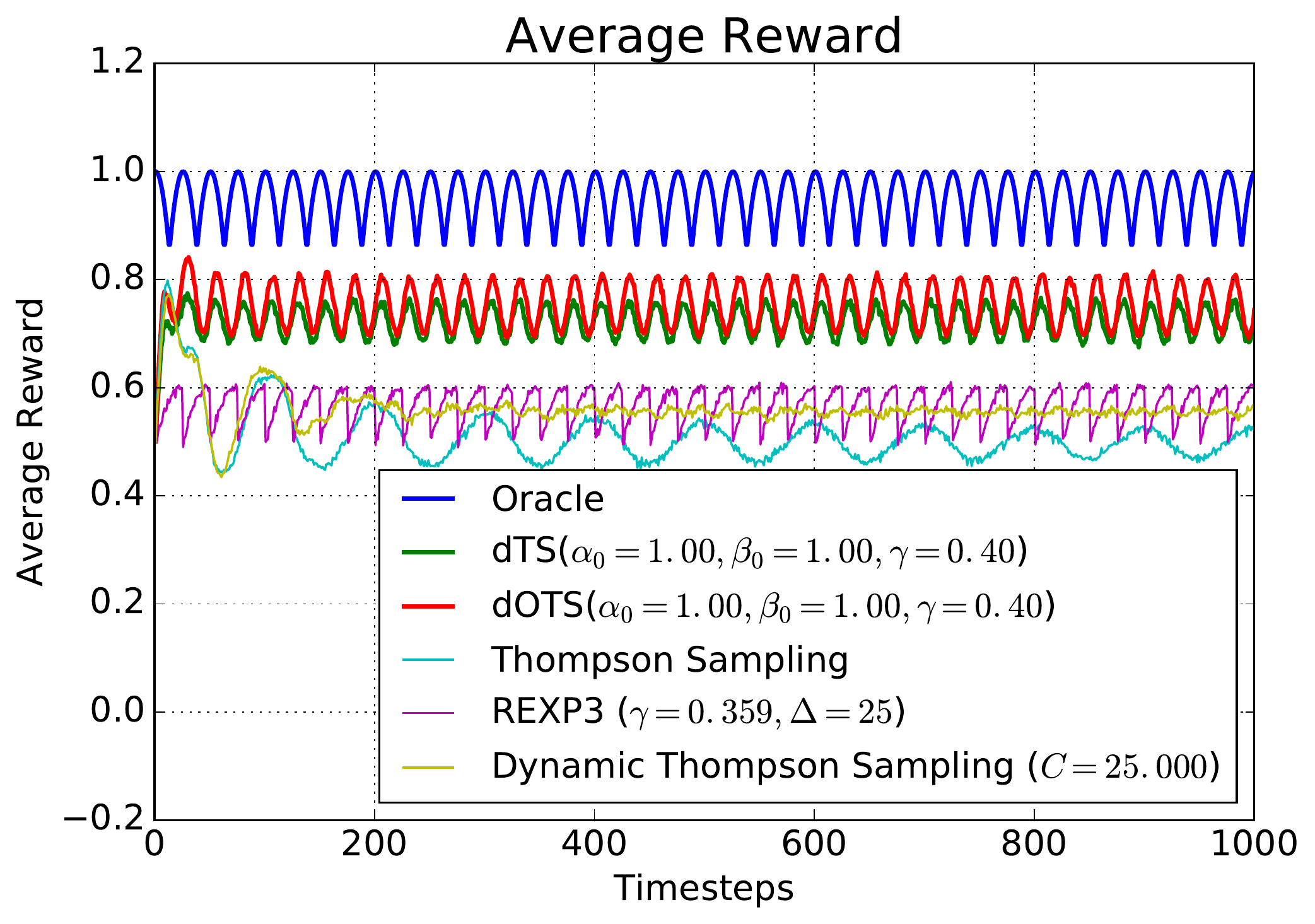}
                \caption{Fast Varying Environment}
                \label{fig:fve_rewards}
            \end{subfigure}
            \begin{subfigure}[h]{0.33\textwidth}
                \includegraphics[width=\linewidth]{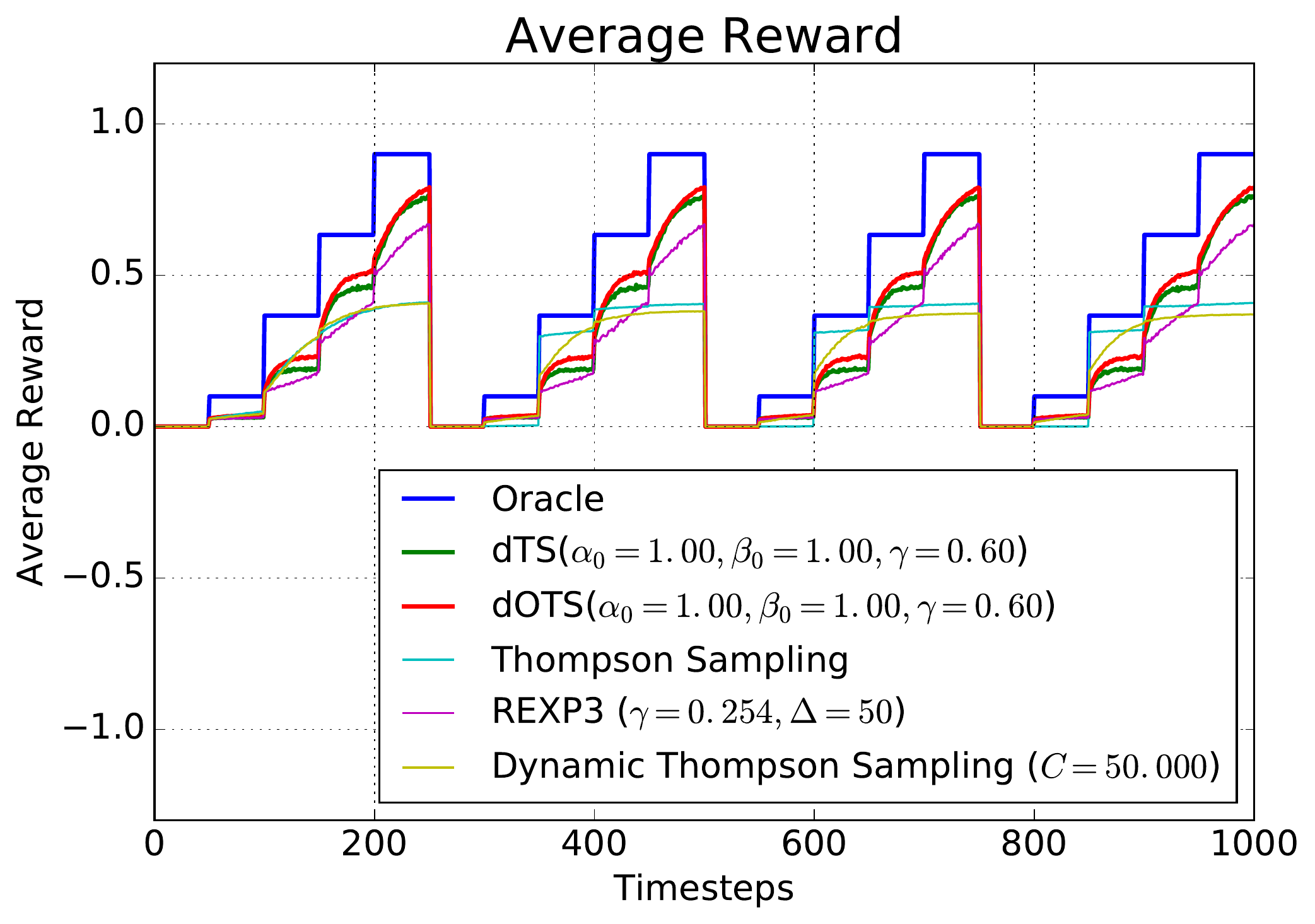}
                \caption{Abruptly Varying Environment}
                \label{fig:ave_rewards}
            \end{subfigure}\\
            \begin{subfigure}[b]{0.33\textwidth}
                \includegraphics[width=\linewidth]{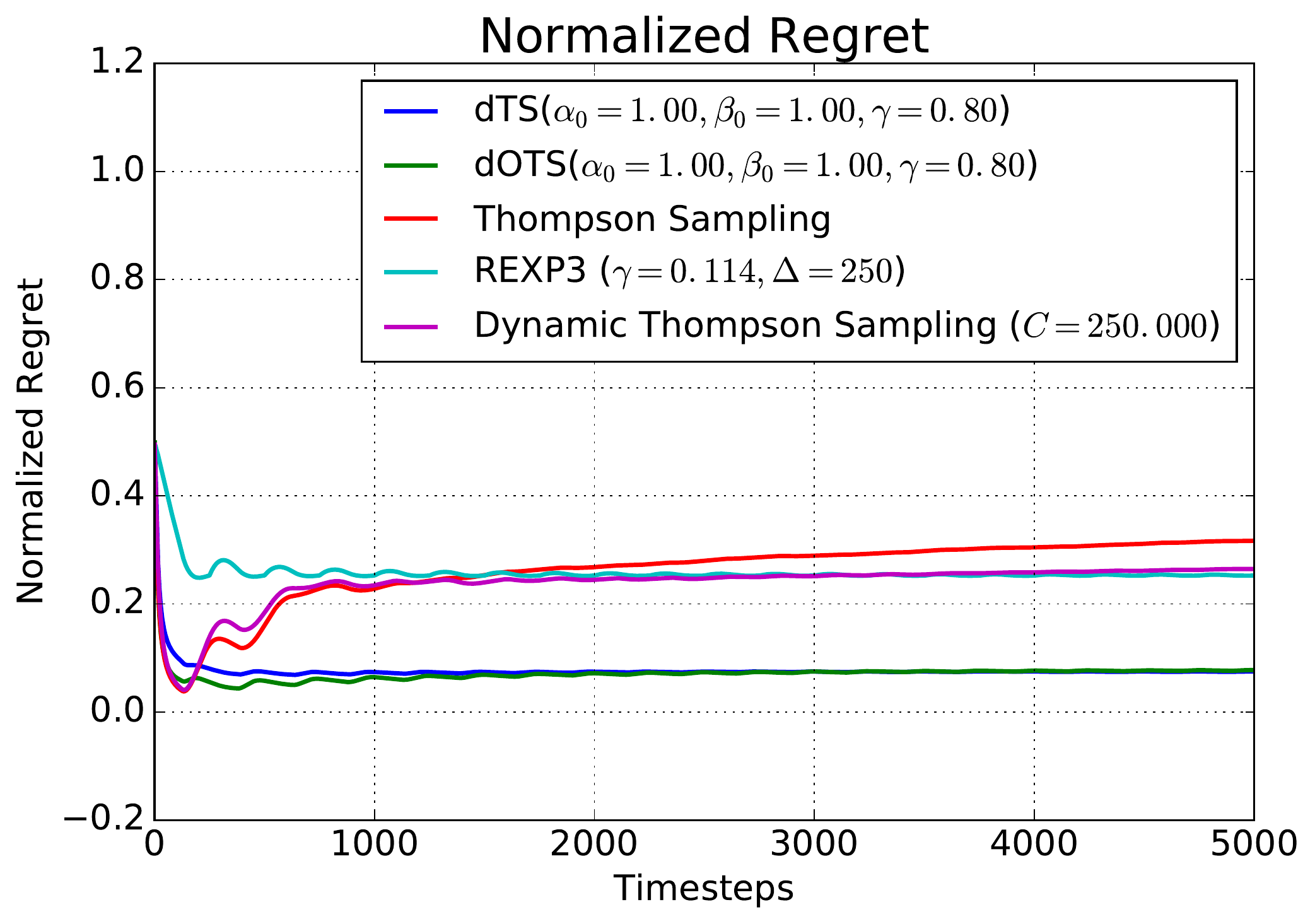}
                \caption{Slow Varying Environment}
                \label{fig:sve_regret}
            \end{subfigure}
            \begin{subfigure}[b]{0.33\textwidth}
                \includegraphics[width=\linewidth]{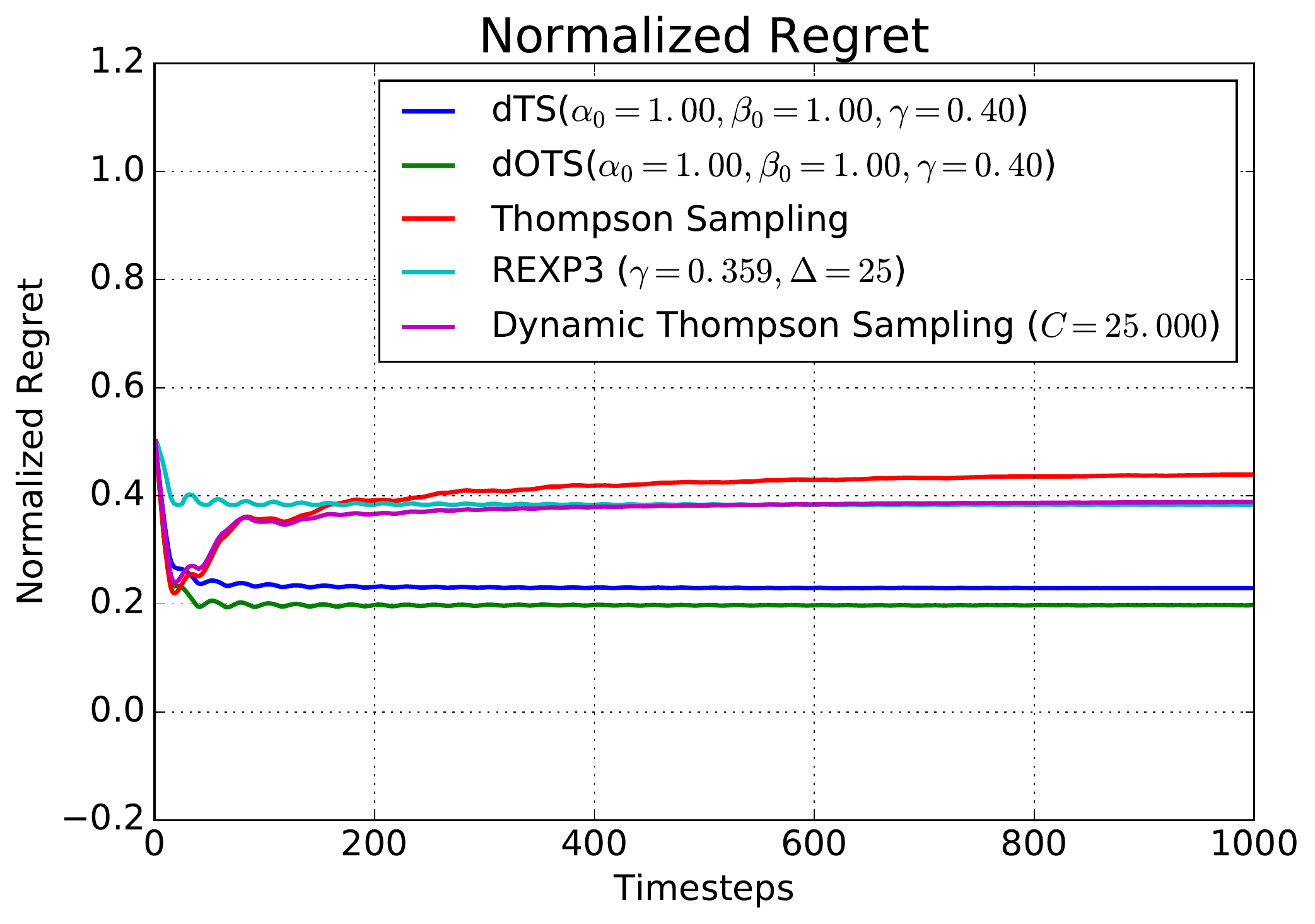}
                \caption{Fast Varying Environment}
                \label{fig:fve_regret}
            \end{subfigure}
            \begin{subfigure}[b]{0.33\textwidth}
                \includegraphics[width=\linewidth]{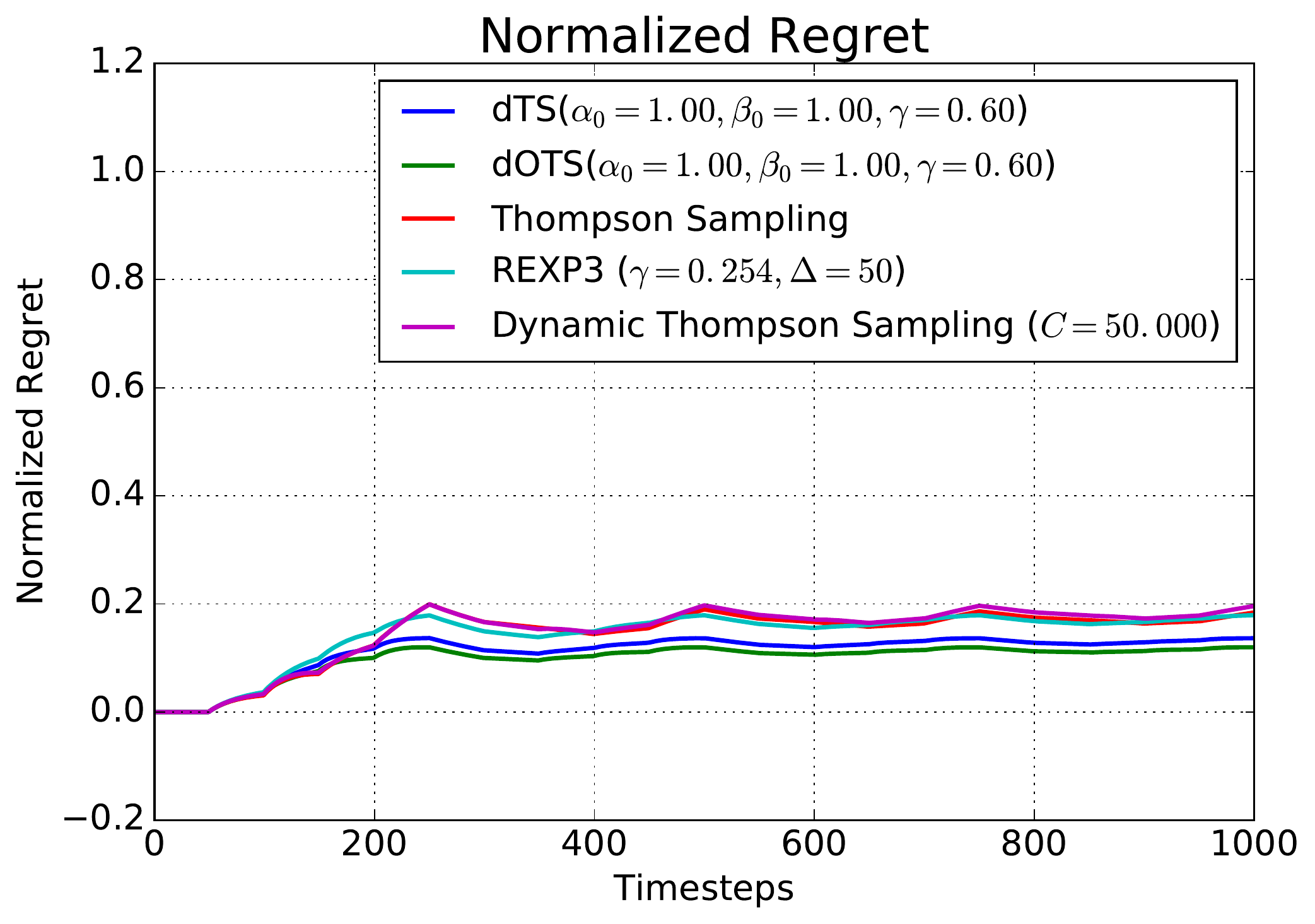}
                \caption{Abruptly Varying Environment}
                \label{fig:ave_regret}
            \end{subfigure}
            \caption{Performance comparison in different scenarios}
            \label{fig:perf_comparison}
        \end{figure}

        To study how regret grows over time, we consider three different environments. These environments are
        similar to the environment discussed in \cite{Besbes2014a}. In all the
        environments, we simulated a four armed Bernoulli bandit with changing success probability. The parameters
        for the comparison algorithms are taken to be the optimal values for each environment as discussed in their
        respective papers. For more information about environment and calculation of parameter values for algorithms, 
        refer supplementary material.
        \\
        \textbf{Slow Varying Environment:} In slow varying environment, we took the success probability of
                each arm as a sinusoidal function in time, limited between $0$ and $1$. To make the environment
                change slowly, the period of all sinusoidals are set as $1000$ timesteps. To have different
                success probabilities at each instant, offset of sinusoidals are taken as $0,\frac{\pi}{2},
                \pi$ and  $\frac{3\pi}{2}$.
                The results are given in 
                Figures \ref{fig:sve_rewards} and \ref{fig:sve_regret}.
        
        \textbf{Fast Varying Environment:} For fast varying environment, the period of the sinusoidal
                discussed above is taken as $100$. The offsets are kept as same as that for slow varying environment.
                Results are given in 
                Figures \ref{fig:fve_rewards} and \ref{fig:fve_regret}.

        \textbf{Abruptly Varying Environment:} For abrupt variations in environment, we assumed the success
            probability of each arm changes abruptly once in a period, going from $0$ to a higher
            value and stays there. We assume an environment where all the $4$ arms starts with $p = 0$.
            Arm-1 changes to $p = 0.10$ at $t = 50$, arm-2 to $p = 0.37$ at $t = 100$, arm-3 to $p = 0.63$
            at $t = 150$ and arm-4 to $p = 0.90$ at $t = 200$. All arms come down to $p = 0$ at $t = 250$ and
            the cycle starts again. Results are given in 
            Figures \ref{fig:ave_rewards} and \ref{fig:ave_regret}.
        
        From the results in 
        Figure \ref{fig:perf_comparison},
        we can observe that the algorithms
        proposed for non-stationary cases - Dynamic TS, REXP3, dTS and dOTS - are able to maintain an almost constant
        normalized regret in various non-stationary environments. However, Thompson Sampling, primarily being an
        algorithm for stationary environments, experiences a growing normalized regret as expected.
        
        From figures \ref{fig:sve_regret} and \ref{fig:fve_regret}, we can see that all algorithms except REXP3 experiences
        a low regret in the initial phase. Then the regret for both TS and Dynamic TS grows over the time while
        dTS and dOTS maintains a low average regret. REXP3, being an algorithm based on the idea of randomized
        exploration, experiences trouble in catching up with the dynamic oracle during initial phase. 
        By the time it gets enough samples to confidently identify the optimal arm, the optimal arm itself changes
        to another
        and hence, again faces trouble in keeping up with the dynamic oracle. 
        However, after sufficient number of timesteps (depending on the environment), it seems to be capable of maintaining
        a constant average regret.
        
        Both dTS and dOTS are able to show a clear learning experience in all the three scenarios. As the variance
        of all the prior distributions keep increasing whenever it is not played, dTS and dOTS do not
        have a trouble in keeping up with dynamic oracle. Even though Dynamic TS also uses similar discounting technique,
        the absence of a mechanism to systematically increase the variance of unplayed arms hurts it during the
        later phase of the game, where it finds itself difficult explore the arms.
        
        Another interesting observation is that, performance of dOTS is better than dTS; even in a fast varying
        environment. This may seem counter-intuitive because by increasing the exploitative value of each arm,
        dOTS tends to pick arms with better empirical mean. However it appears that this approach is helping
        dOTS to take better decisions, even in abruptly changing environments.
        
        An extensive study of comparison against various algorithms is provided in the supplementary
        material.

     \subsection{Increasing the number of arms}
\begin{figure}[h]
             \begin{subfigure}[b]{0.33\textwidth}
                 \includegraphics[width=\linewidth]{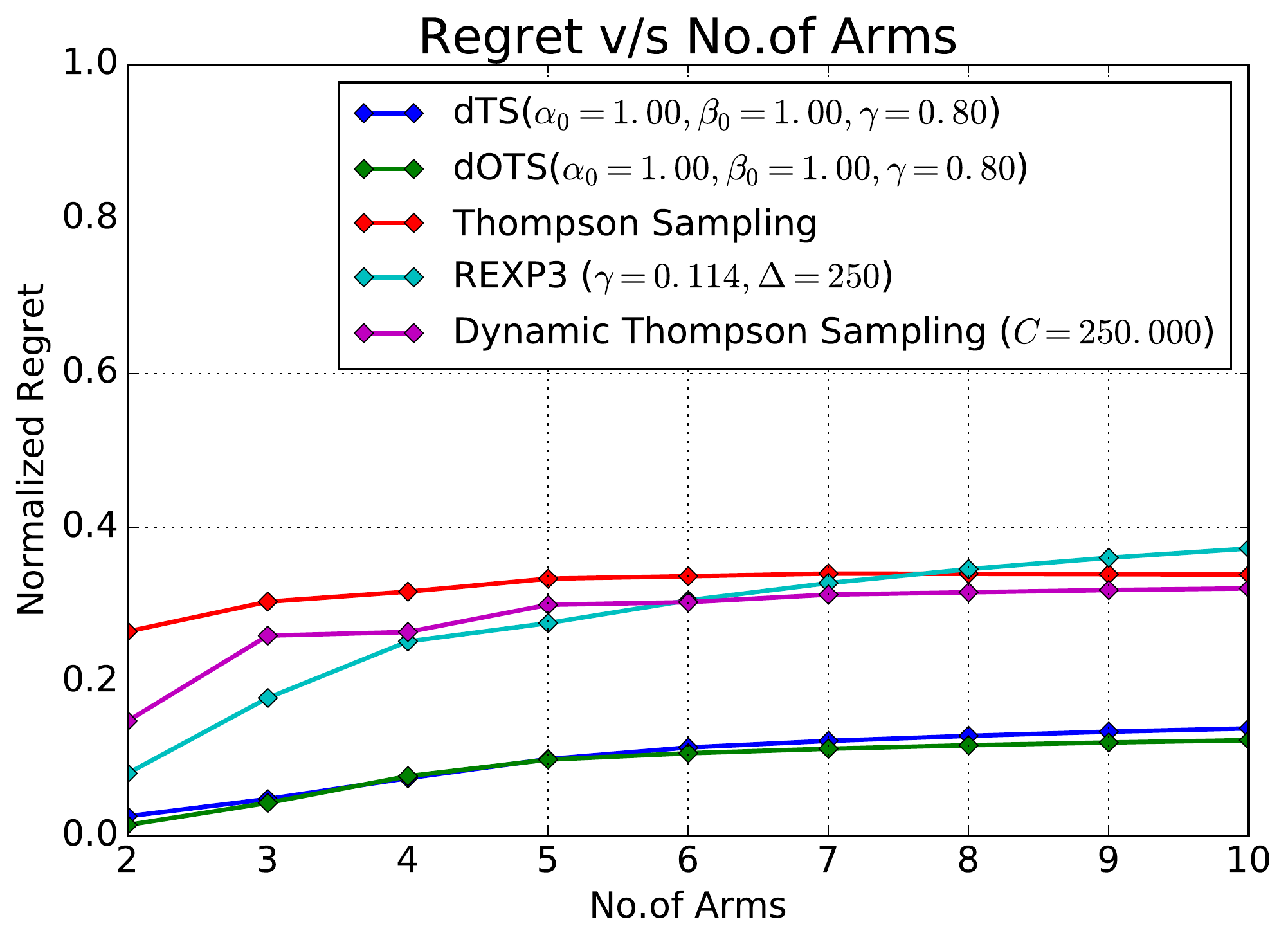}
                 \label{fig:arms_sve}
                 \caption{Slow Varying Environment}
             \end{subfigure}
             \begin{subfigure}[b]{0.33\textwidth}
                 \includegraphics[width=\linewidth]{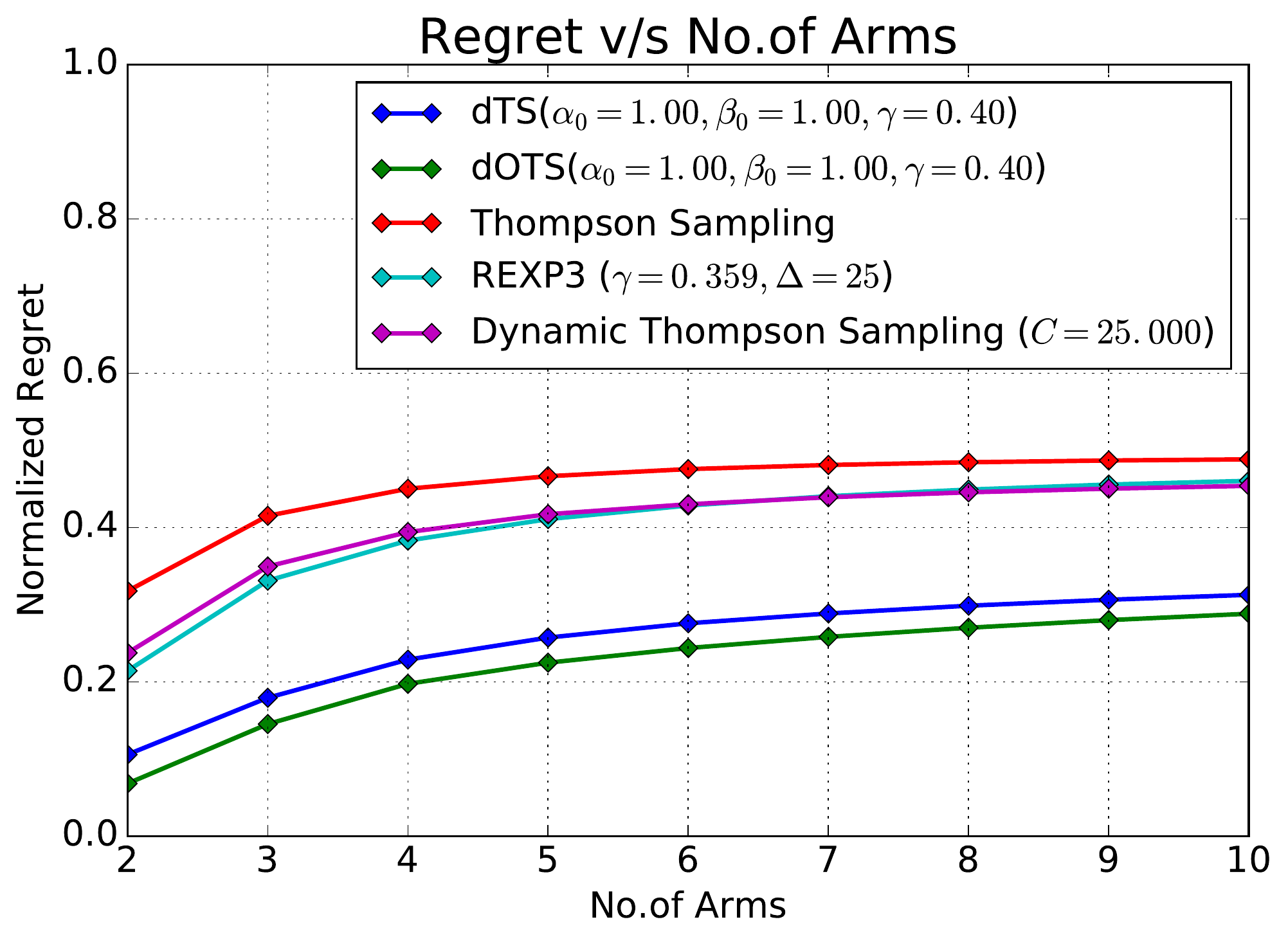}
                 \label{fig:arms_fve}
                 \caption{Fast Varying Environment}
             \end{subfigure}
             \begin{subfigure}[b]{0.33\textwidth}
                 \includegraphics[width=\linewidth]{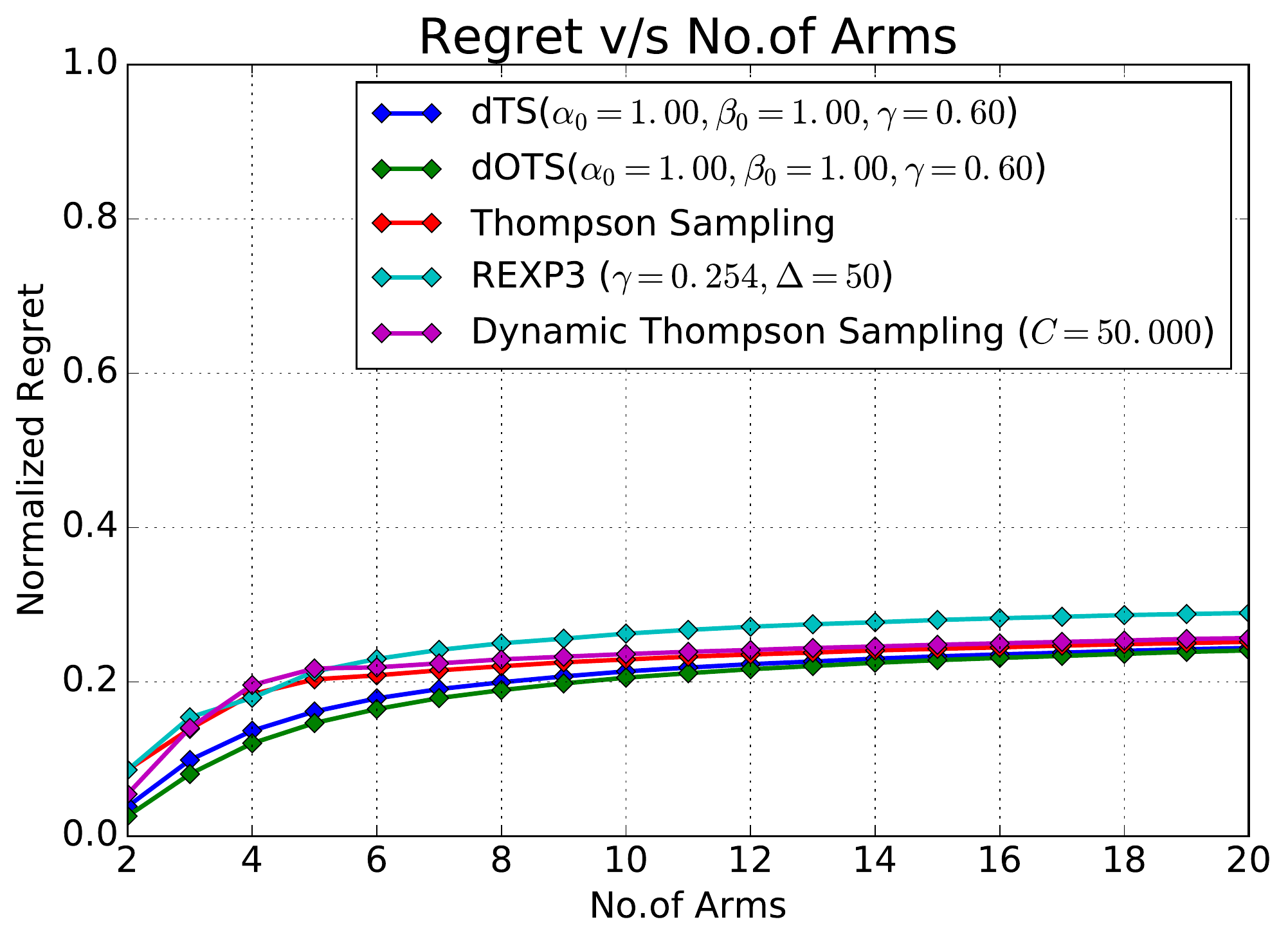}
                 \label{fig:arms_ave}
                 \caption{Abruptly Varying Environment}
             \end{subfigure}
             \caption{Effect of increasing the number of arms}\label{fig:arms_effect}
         \end{figure}

         Figure \ref{fig:arms_effect} shows how normalized regret behaves as the number of arms in the
         bandit increases. Bandit environment for this experiment is similar to the one described in 
         previous section, except for the number of arms. For sinusoidal environments, the offset of each arm
         is set at equal intervals between $0$ and $2\pi$. For abrupt changes, the change points are distributed
         equally in the period for each arm. The results are provided for a run of $5000$ timesteps.
        
        From the figure, we can observe that both dTS and dOTS are able to maintain a good margin of
        regret in both
        slow and fast varying environments. But in abruptly changing environment, both the algorithms
        eventually reaches the same regret as comparison algorithms, all of them growing with increase
        in number of arms. Again, Thompson Sampling is showing a high regret in both slow and 
        fast varying environments.
        A key point to note is that the parameter values of the none of the algorithms are recalculated
        for the increased number of arms. Hence, this experiment indirectly shows how strongly
        the values of parameters depend on the number of arms.
        
        As the number of arms increases, REXP3 shows an increased regret in slow varying environment.
        This could be because of the strong dependence of parameters of REXP3 on the number of arms. In
        abruptly varying environment, this effect is more profound and REXP3 experiences the maximum
        regret, even more that that of TS. These experiments also show that the parameter value of
        dTS and dOTS may not be strongly dependent on the number of arms.

\section{Concluding Remarks}
    In this paper, we proposed a Bayesian algorithm for non-stationary bandits. Based
    on the idea of systematically increasing the variance of prior distribution of unplayed arms and
    utilizing exponential filtering to \emph{forget} the past observations, Discounted Thompson Sampling (dTS) and 
    Discounted Optimistic Thompson Sampling (dOTS) algorithms are able to perform better in different worst case 
    scenarios. We also provided the exact expression
    for the probability of picking a sub-optimal arm in a two armed bandit setting. To the best of our knowledge, this
    is the first time this probability is discussed for non-integer parameters of Thompson Sampling. 
    By providing a general expression for the probability of a sub-optimal arm being picked, we believe
    that this work will help in analyzing the popular TS algorithm in a wide range of scenarios.
    Future work can include analyzing the performance of the proposed algorithms and
    bounding the regret incurred during the game.

\small
\bibliographystyle{unsrt}
\bibliography{library}

\section*{Appendix A: Hypergeometric Functions}
    Gauss Hypergeometric function is defined as
    \begin{align}
        \textstyle\pFq{2}{1}{a_1,a_2}{b_1}{z} &= \sum \limits_{k=0}^{\infty} \frac{(a_1)_k (a_2)_k }{(b_1)_k} \frac{z^{k}}{k!}
            & |z| < 1 \lor |z| = 1 \land Re(b_1-a_1-a_2) > 0,
    \end{align}
    where $(q)_{n}$ is the Pochhammer symbol defined as
    \begin{align}
        (q)_{n} &= \frac{\Gamma(q+n)}{\Gamma(q)}.
    \end{align}
    Here, $\Gamma(\cdot)$ is the Gamma function.
    
    By substituting $a_1 = S_{1,t} + \alpha_0 + S_{2,t}+ \alpha_0$, $a_2 = 1-(F_{1,t}+\beta_0)$,
    and $b_1 = S_{1,t} + \alpha_0 + S_{2,t} + \alpha_0 + F_{2,t} + \beta_0$ in the condition
    $b_1 - a_1 - a_2 > 0$, we get
    \begin{align*}
        \beta_0 &> \frac{1-(F_{1,t}+F_{2,t})}{2}
    \end{align*}
    Taking $F_{k,t} = 0$, we need $\beta_0 > \frac{1}{2}$. Hence, for the probability expression to hold at all
    times, the $\beta$ parameter should be greater than $\frac{1}{2}$ for Beta prior.
    
    For $\textstyle\pFq{3}{2}{a_1,a_2,a_3}{b_1,b_2}{z}$, we have
    \begin{align}
        \textstyle\pFq{3}{2}{a_1,a_2,a_3}{b_1,b_2}{z} &=
            \sum \limits_{k=0}^{\infty} \frac{(a_1)_{k} (a_2)_{k} (a_3)_{k}}{(b_1)_{k} (b_2)_{k}} \frac{z^{k}}{k!}
            & |z| < 1 \lor
                |z| = 1 \land Re\left( \sum\limits_{j=1}^{2} b_j - \sum \limits_{j=1}^{3} a_j\right) > 0
    \end{align}
    From the conditions, we will get $ F_{1,t} + \beta_0 + F_{2,t} + \beta_0 > 0$. By taking $F_{k,t} = 0$,
    we need $\beta_0 > 0$, which is always true.
\section*{Appendix B: Comparison with state-of-the-art algorithms}
    This section includes the results of extensive simulation studies we did to evaluate dTS and dOTS. First,
    we introduce the environment used for the simulation studies. Then, we study the impact of $gamma$ parameter
    in the regret of the algorithm. Later in this section, we provide results of dTS and dOTS compared with 
    five state-of-the-art algorithms - REXP3, Dynamic TS, Discounted-UCB, Sliding Window UCB and EXP3-IX -
    for non-stationary bandits.
    \subsection*{Environment}
\begin{figure}[!h]
            \begin{subfigure}[b]{0.33\textwidth}
                \includegraphics[width=\linewidth]{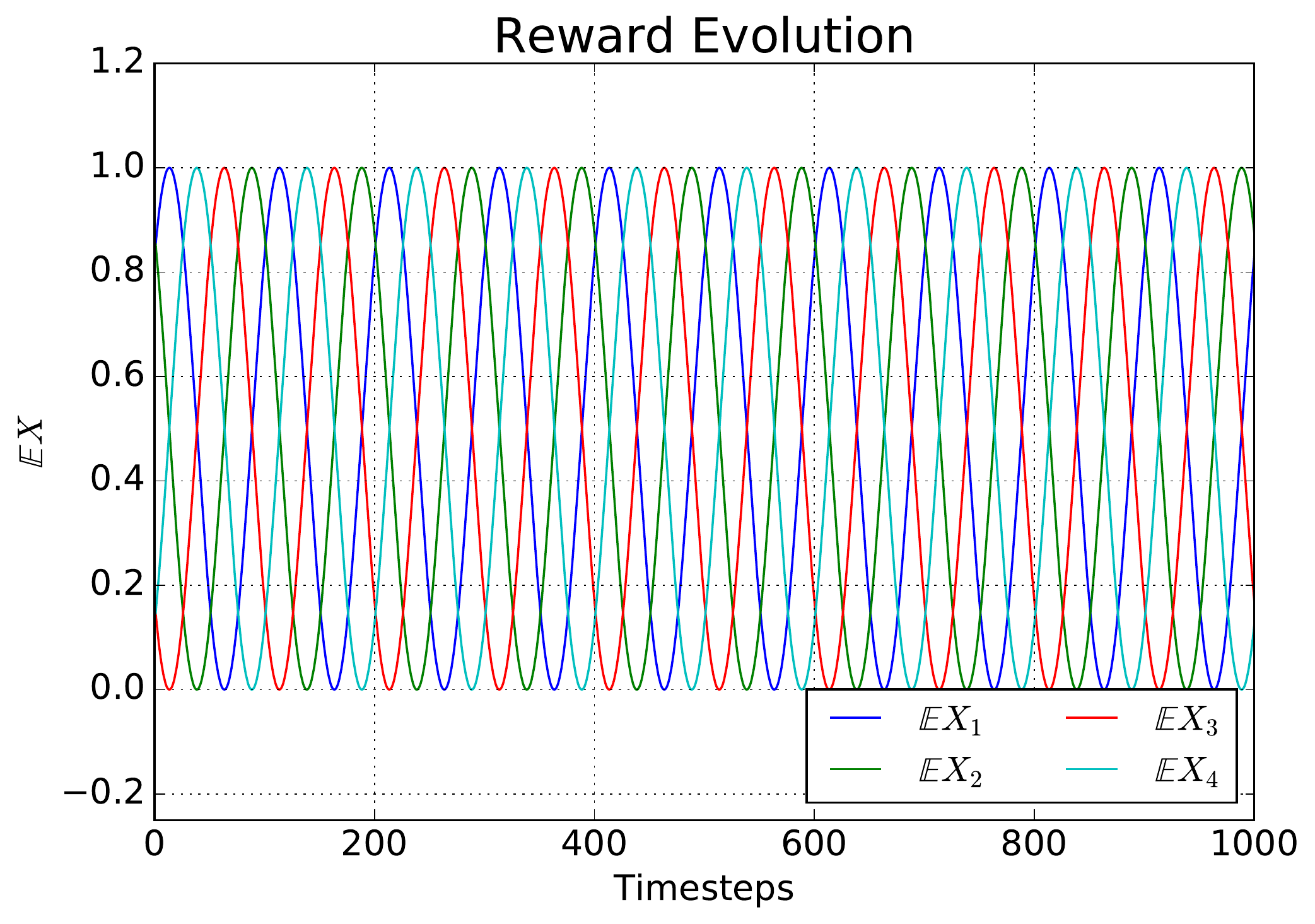}
                \caption{Fast varying environment}
                \label{fig:env_fve}
            \end{subfigure}
            \begin{subfigure}[b]{0.33\textwidth}
                \includegraphics[width=\linewidth]{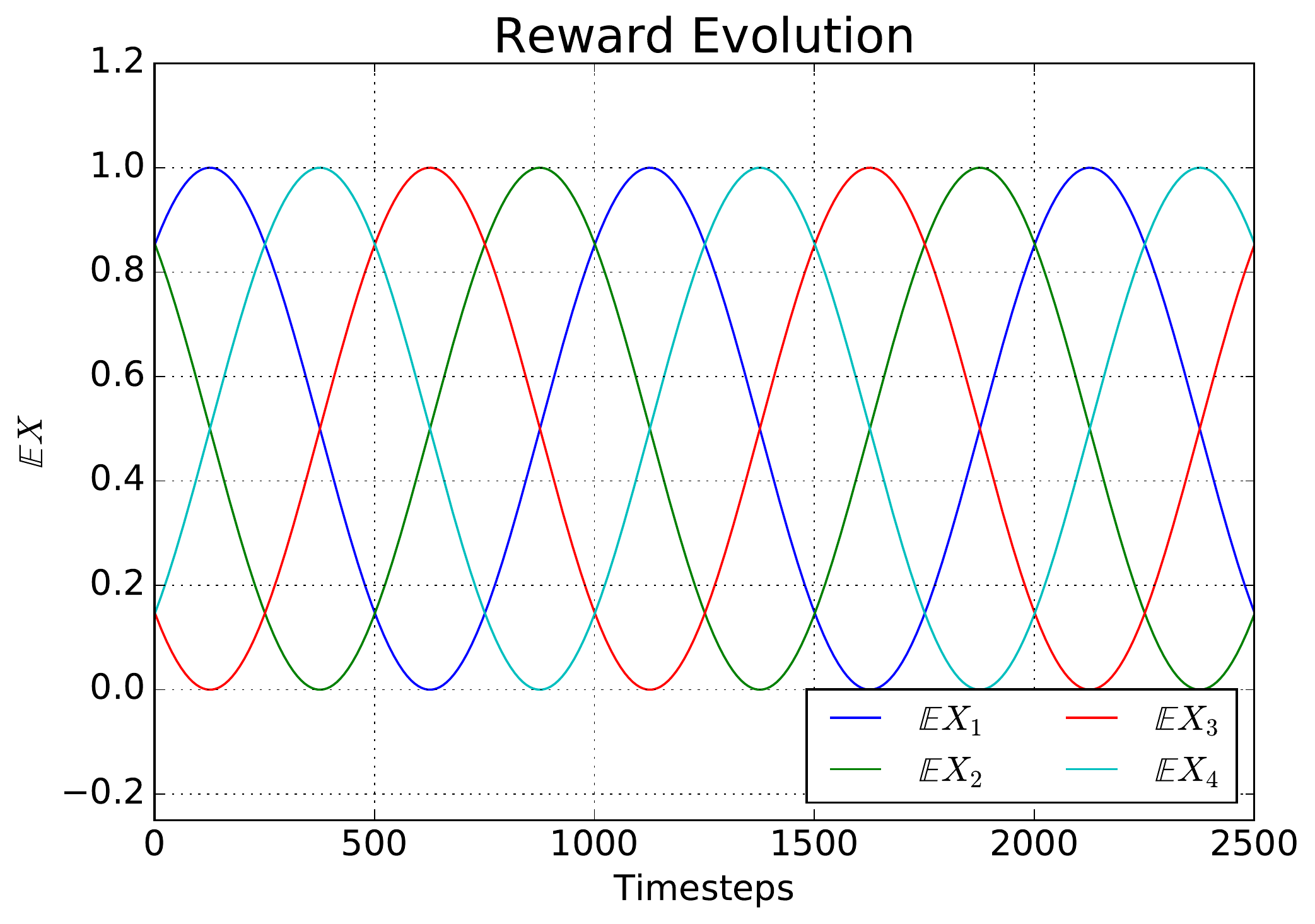}
                \caption{Slow varying environment}
                \label{fig:env_sve}
            \end{subfigure}
            \begin{subfigure}[b]{0.33\textwidth}
                \includegraphics[width=\linewidth]{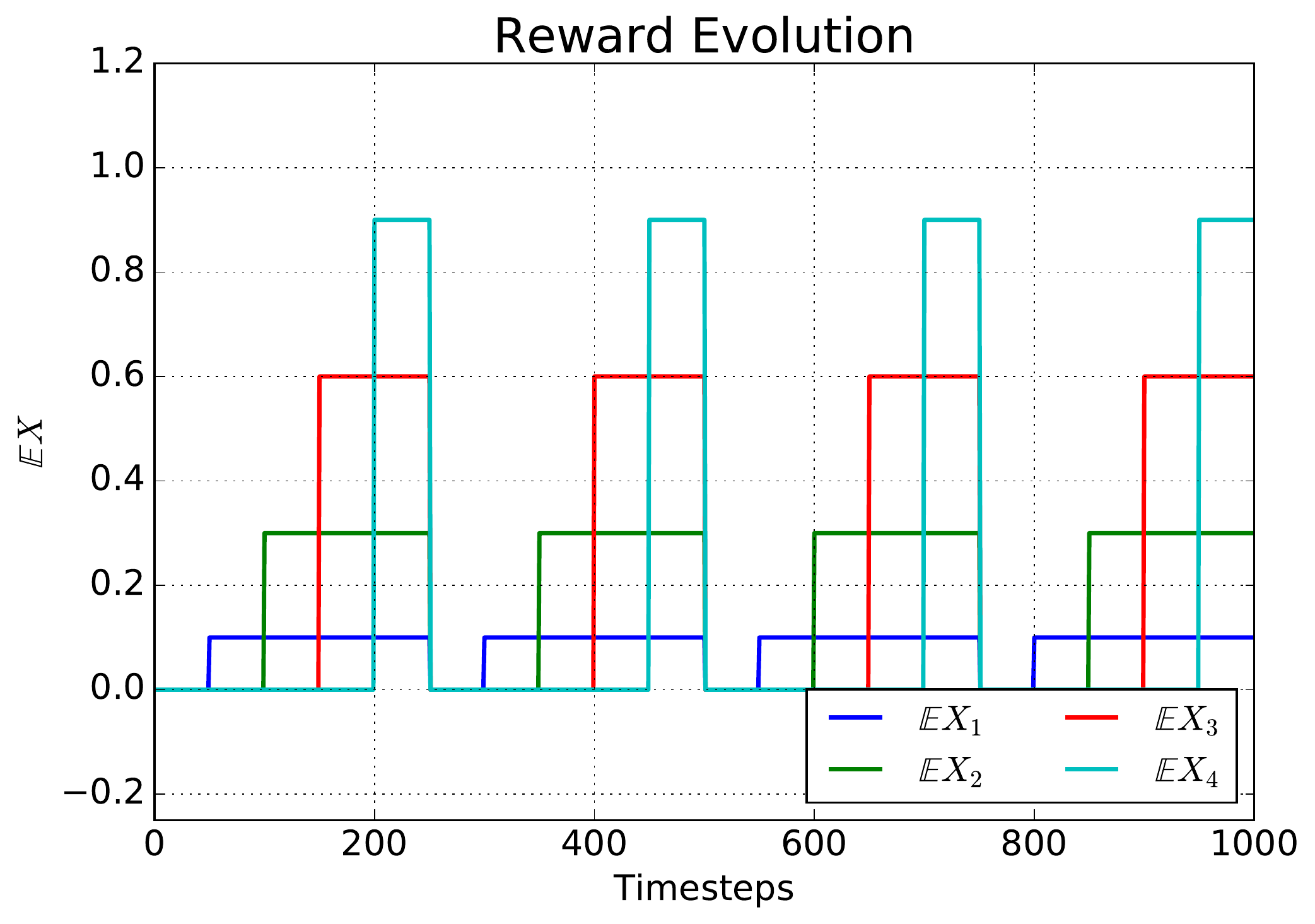}
                \caption{Abruptly varying environment}
                \label{fig:env_ave}
            \end{subfigure}
            \caption{Different Environments used for evaluation of algorithms}
            \label{fig:env_rewards}
        \end{figure}

        For comparing against various state of the art algorithms, the following environments are used:
        \begin{enumerate}
            \item Fast Varying Environment: To simulate a fast varying environment, we consider a bandit with four Bernoulli 
                    arms where the expected rewards vary in a sinusoidal fashion with a period of 100 time steps. Each arm
                    has an offset which is an integer multiple of $\pi/2$ to create variation.
            \item Slow Varying Environment: For slow varying environment, a bandit similar to the one mentioned above is
                    taken, but the period of the sinusoidal wave is taken as 1000 time steps.
            \item Abruptly Varying Environment: For abruptly changing bandit, we designed a four armed Bernoulli bandit with
                    the expected reward of each arm abruptly going from 0 to some finite value at random time and it stays at
                    new value till the end of period. The optimal arm switches every $50$ time steps
                     and this entire procedure is repeated every $250$ time steps as shown in Figure \ref{fig:env_ave}.
        \end{enumerate}  
        
        All the results provided below are averaged over $1000$ independent runs. Also, unless stated otherwise, all
        the experiments are run for a time horizon, $T = 5000$.
        
    \subsection*{Effect of $\gamma$}
        The discounting factor introduced in the proposed algorithm for the trade-off between \textit{remembering} and
        \textit{forgetting} plays an important role in the performance of the algorithm. A high value of $\gamma$ will
        make the algorithm remember the past rewards for more time and a low value of $\gamma$ will make the algorithm
        forget past rewards faster. Hence, setting the value of $\gamma$ has an impact on the performance of algorithm.
        In this section, we study the effect of setting different values for $\gamma$ and try to analyse its impact.

        \begin{figure}[h]
            \begin{subfigure}[b]{0.33\textwidth}
                \includegraphics[width=\linewidth]{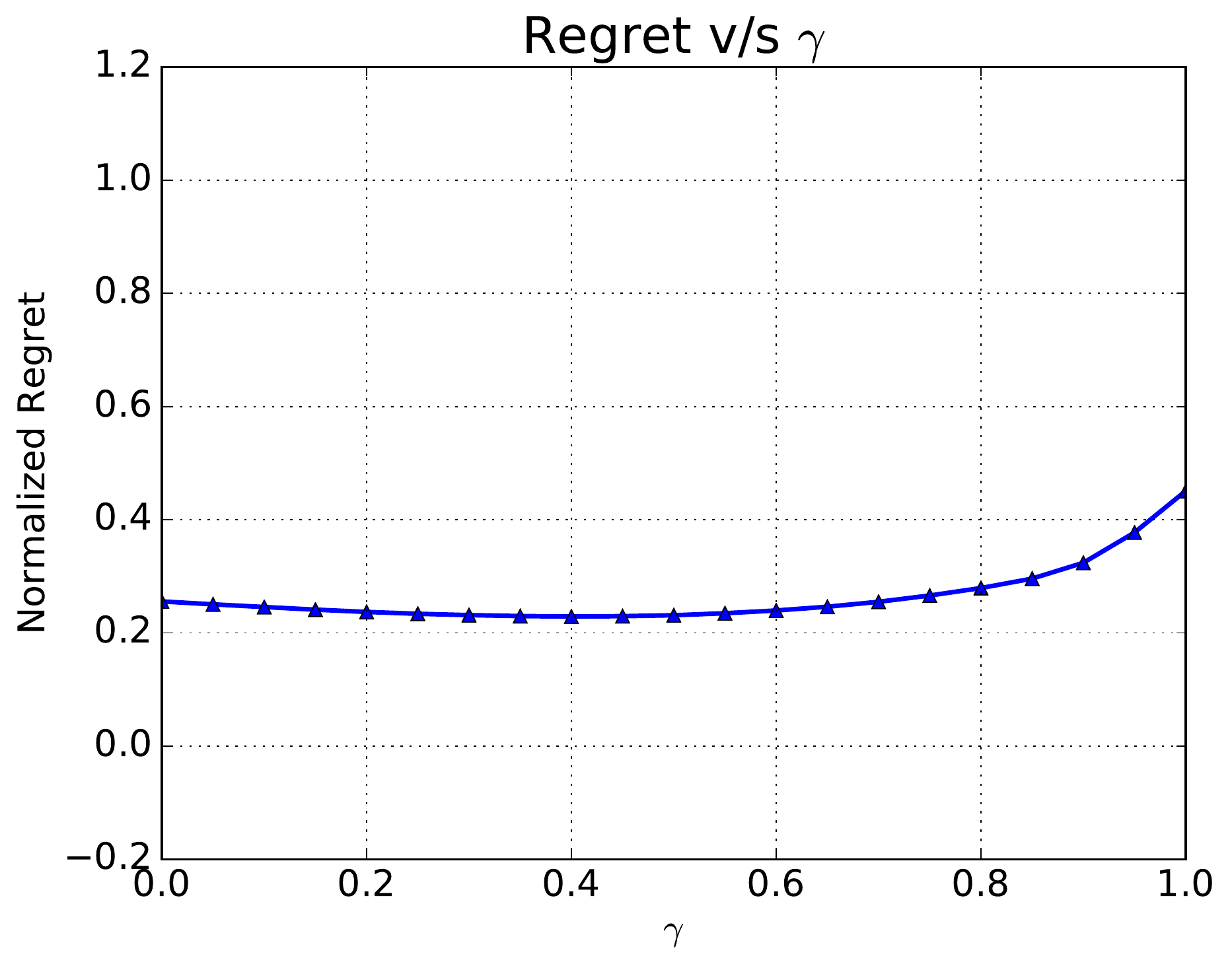}
                \caption{Fast varying environment}
                \label{fig:gamma_dTS_fve}
            \end{subfigure}
            \begin{subfigure}[b]{0.33\textwidth}
                \includegraphics[width=\linewidth]{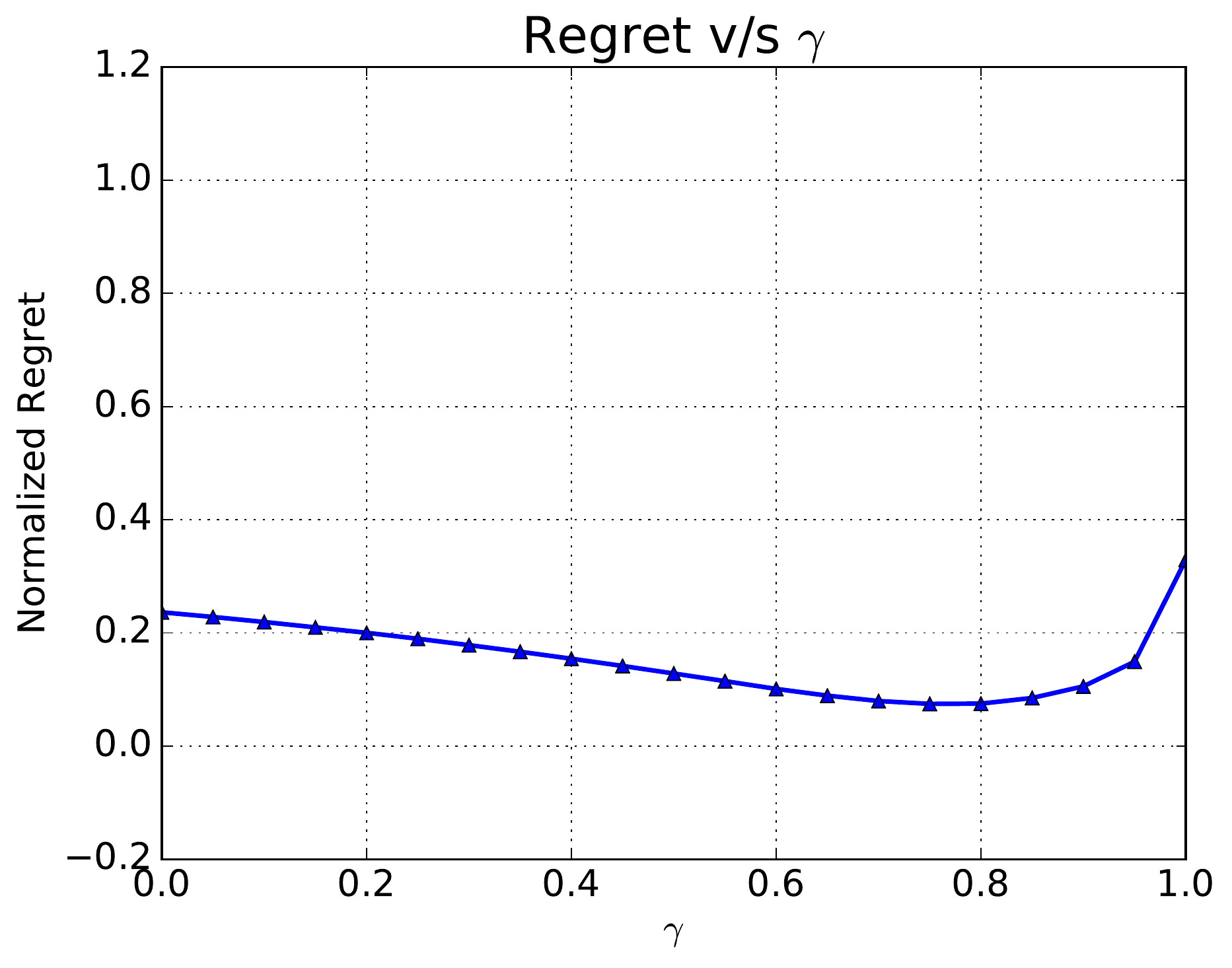}
                \caption{Slow varying environment}
                \label{fig:gamma_dTS_sve}
            \end{subfigure}
            \begin{subfigure}[b]{0.33\textwidth}
                \includegraphics[width=\linewidth]{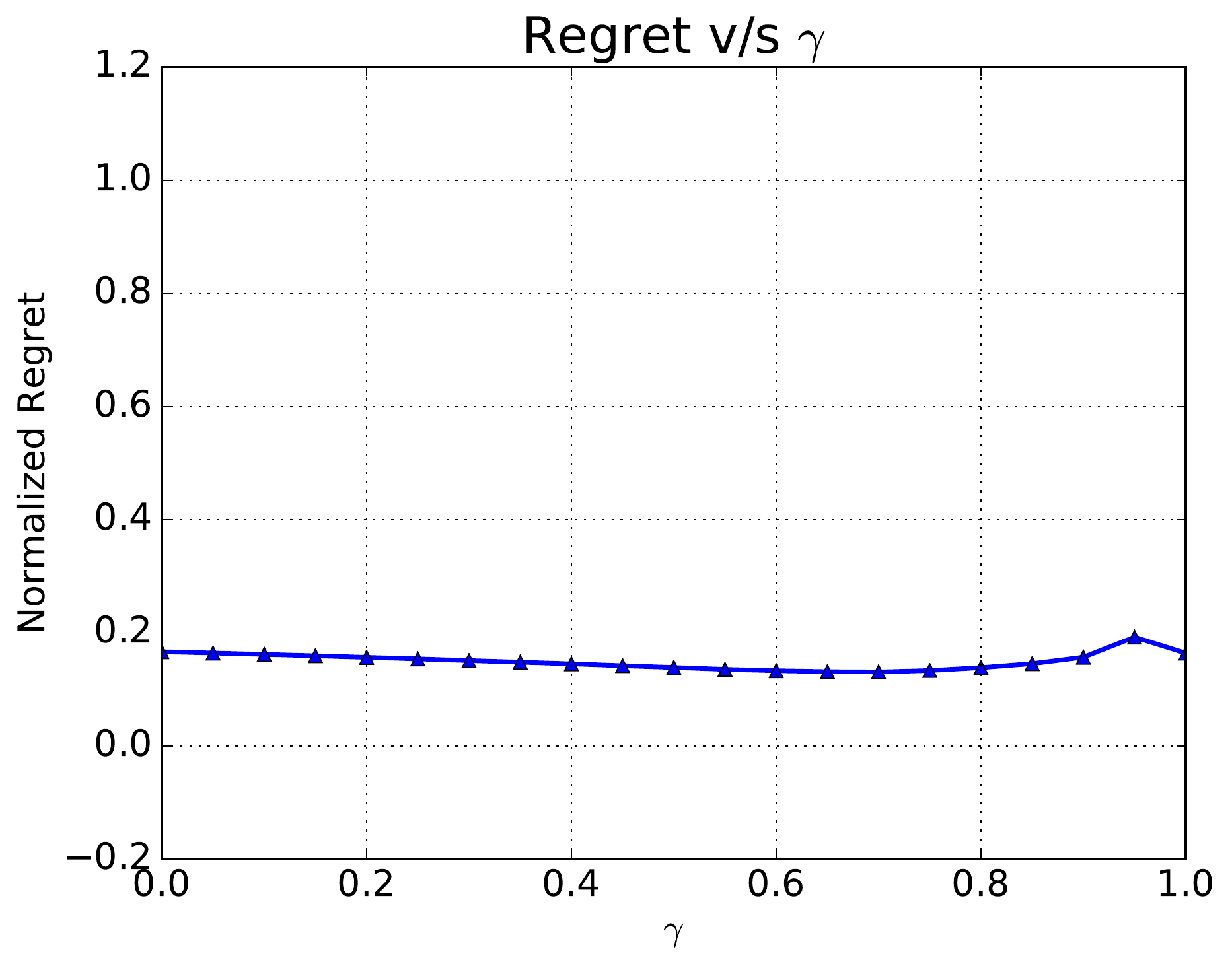}
                \caption{Abruptly changing environment}
                \label{fig:gamma_dTS_ave}
            \end{subfigure}
            \caption{Effect of $\gamma$ in regret with parameters $\alpha_0 = 1, \beta_0 = 1$ on dTS}
            \label{fig:gamma_dTS}
        \end{figure}
    
        \begin{figure}[h]
            \begin{subfigure}[b]{0.33\textwidth}
                \includegraphics[width=\linewidth]{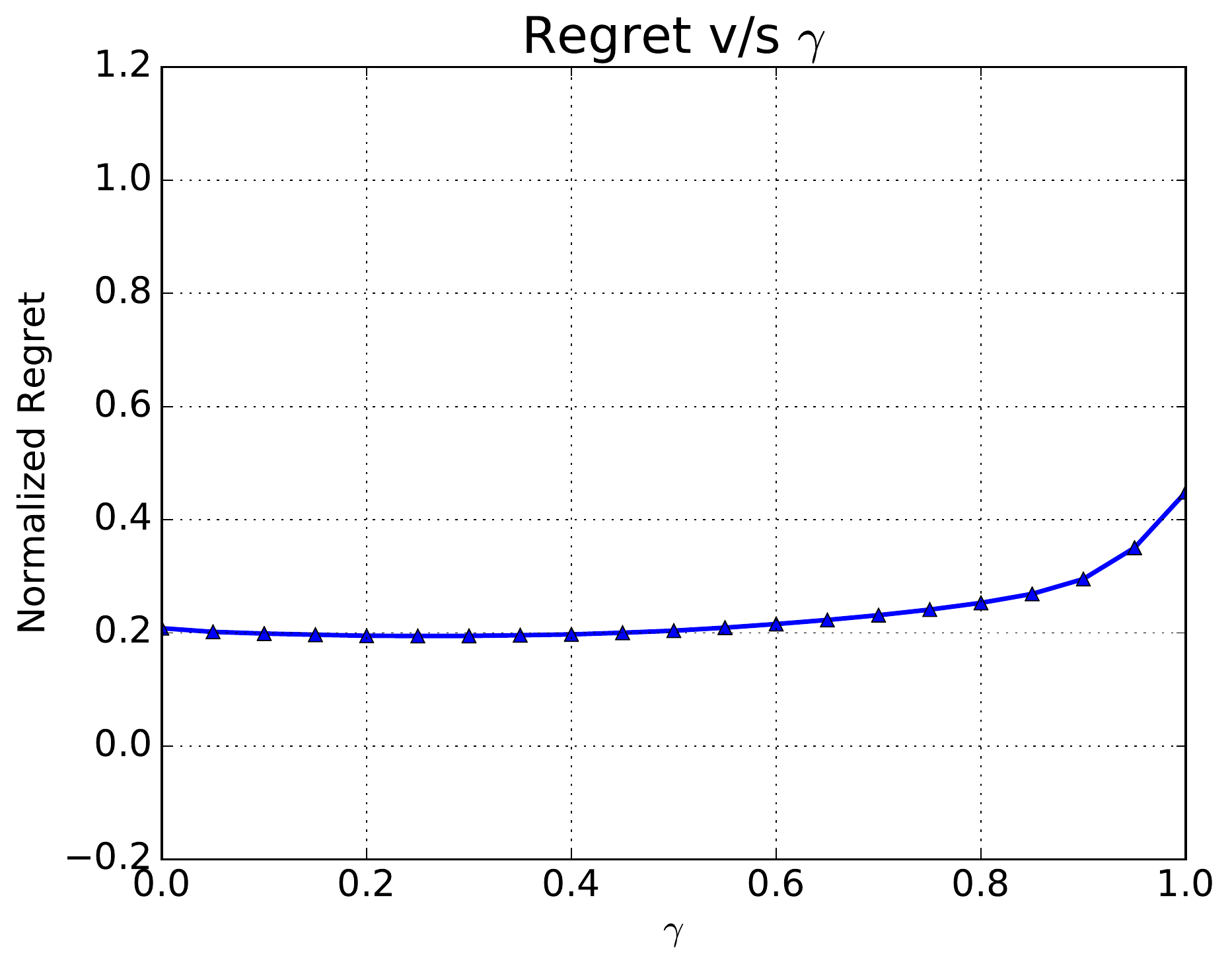}
                \caption{Fast varying environment}
                \label{fig:gamma_dOTS_fve}
            \end{subfigure}
            \begin{subfigure}[b]{0.33\textwidth}
                \includegraphics[width=\linewidth]{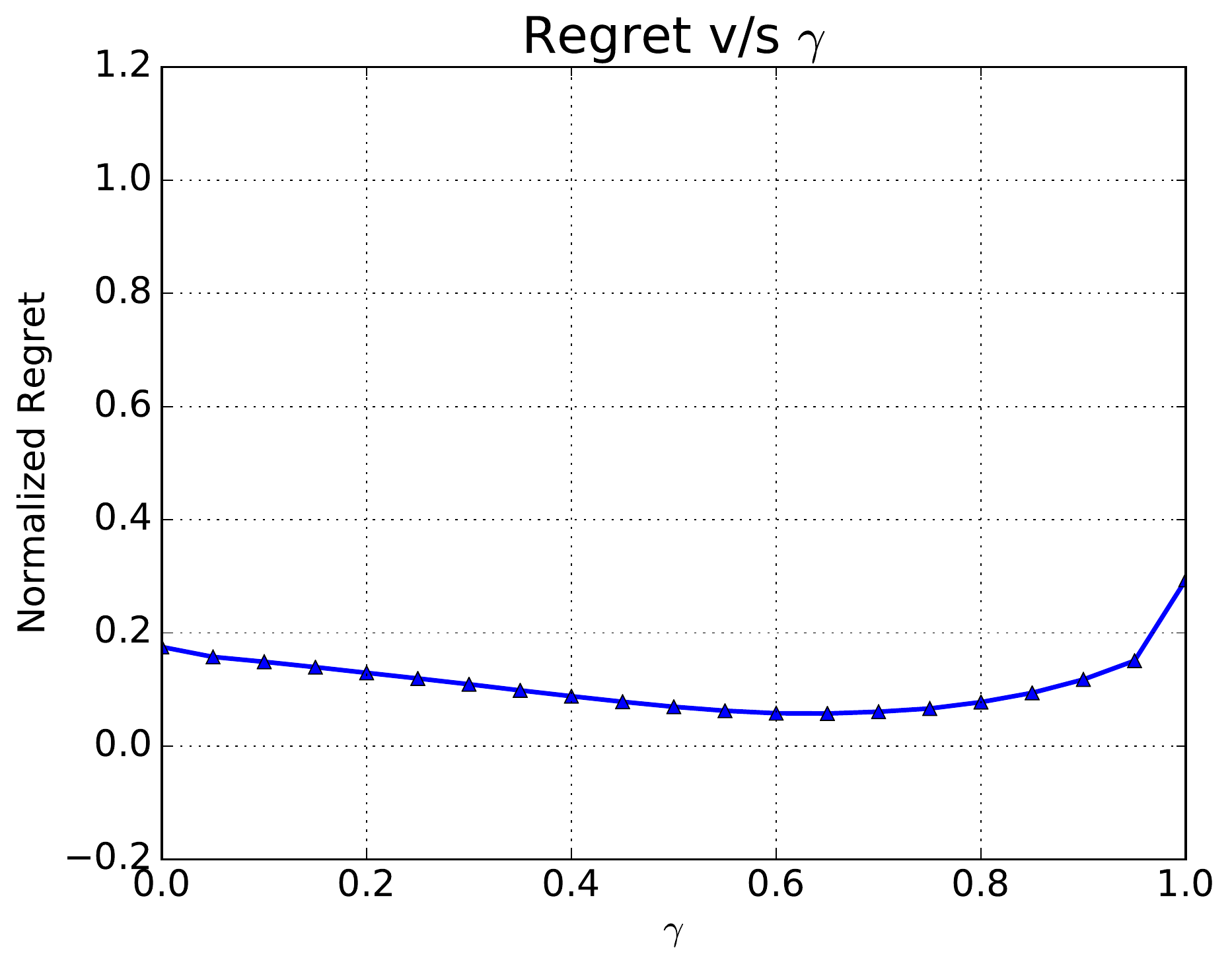}
                \caption{Slow varying environment}
                \label{fig:gamma_dOTS_sve}
            \end{subfigure}
            \begin{subfigure}[b]{0.33\textwidth}
                \includegraphics[width=\linewidth]{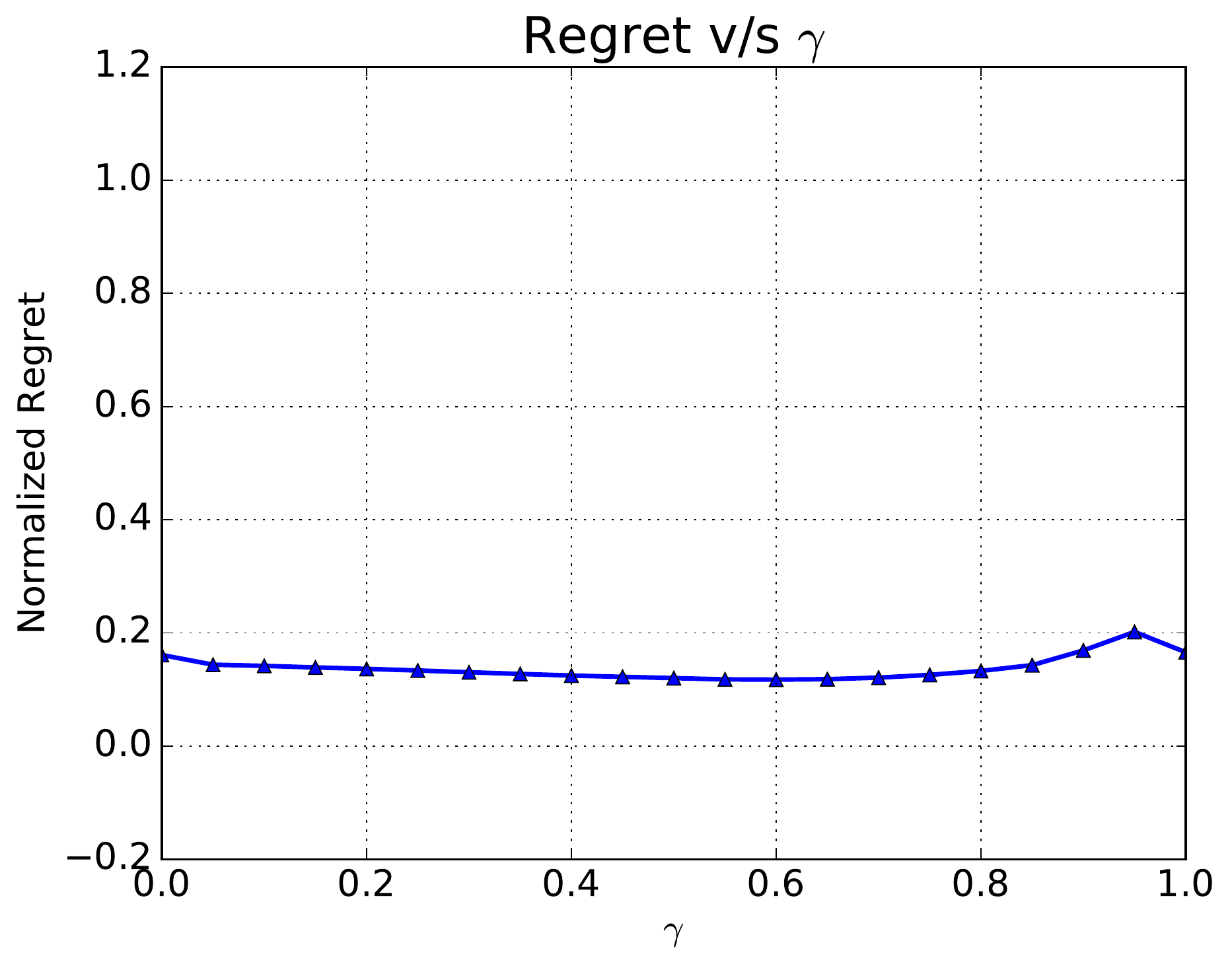}
                \caption{Abruptly changing environment}
                \label{fig:gamma_dOTS_ave}
            \end{subfigure}
            \caption{Effect of $\gamma$ in regret with parameters $\alpha_0 = 1, \beta_0 = 1$ on dOTS}
            \label{fig:gamma_dOTS}
        \end{figure}
        
        Figure \ref{fig:gamma_dTS} shows the variation in regret of dTS for different values of $\gamma$. Figure 
        \ref{fig:gamma_dOTS} shows the same for dOTS. Both the experiments were run for a horizon length of $5000$
        time steps. 

        From both the results, we can see that the regret behaves as a smooth function of $gamma$ in the case
        of dTS and dOTS for slow and fast varying environments. But in case of abruptly varying environment
        (in the special environment we mentioned), regret peaks at $\gamma = 0.95$, which is a surprise.
        With $\gamma = 1.0$, dTS (and dOTS) acts like TS (and OTS), which is an algorithm for stationary
        bandit case. But the results here show that, at $\gamma = 0.95$, both dTS and dOTS have difficulty
        in forgetting the past.
        
    \subsection*{REXP3}
        REXP3 is proposed in \cite{Besbes2014a} as a near optimal policy for non-stationary bandits. The algorithm has
        two parameters - $\gamma \in [0,1]$, the egalitarianism factor and $\Delta_T$, the time duration for which
        one arm will stay as optimal arm.
        
        For sinusoidal environments, the optimal arm switches every $T/4$ timesteps, where $T$ is the period of sinusoidal
        wave. Hence $\Delta_T = T/4$ is set as the resetting period of REXP3 algorithm. For setting the $\gamma$ parameter,
        the following formula is used:
        \begin{align}
            \gamma &= \min \left\{ 1, \sqrt{\frac{K \log K}{(e-1)\Delta_T}} \right\}        
        \end{align}
        where $K$ is the number of arms.
        Actual values used for simulation are listed in Table \ref{tab:rexp3_params}.
        \begin{table}[H]
            \caption{Comparison with REXP3: Parameters}
            \label{tab:rexp3_params}
            \centering
            \begin{tabular}{lrrrrr}
                \toprule
                                  & Period & \multicolumn{2}{c}{REXP3} & dTS      & dOTS       \\
                Environment       & $T$    & $\Delta_T$&  $\gamma$     & $\gamma$ & $\gamma$   \\
                \midrule
                Fast Varying      &   100  &      25   &  0.3593       & 0.40     & 0.40    \\
                Slow Varying      &  1000  &     250   &  0.1136       & 0.75     & 0.75    \\
                Abruptly Varying  &   250  &      25   &  0.5000       & 0.60     & 0.60    \\
                \bottomrule
            \end{tabular}
        \end{table}
        Results are provided in Section 5.1 of main script.
    
    \subsection*{Dynamic Thompson Sampling}
        Dynamic Thompson Sampling(DTS) proposed in \cite{Gupta2011a} uses the idea of exponential filtering technique to
        adapt to the changes in the environment. One key difference of DTS with the proposed algorithm is the way the
        discounting is applied. DTS applies discounting only to the arm it picks to play and also after a particular
        threshold is crossed($C$). Dynamic Thompson Sampling takes a simple parameter, $C$, which decides when to apply
        discounting ($\alpha_k + \beta_k > C$) and then how much to discount past rewards (by a factor of $\frac{C}{C+1}$).
        
        As there is no analysis of Dynamic Thompson Sampling available, we used to following heuristic argument to set
        the value for $C$ in the experiments. Parameter $C$ starts affecting the algorithm only after $\alpha_k + \beta_k
        > C$. If we know when the change points are occurring in the environment, we can set the value of $C$ to be
        equal to the time interval between change points. Hence for sinusoidal environments, $C$ can be set to $T/4$
        where $T$ is the period of the sinusoidal. For abruptly changing environment, $C$ is taken as the minimum interval
        between two change points in the environment. Values used in simulation are listed in Table \ref{tab:DTS_params}.
        \begin{table}[H]
            \caption{Comparison with Dynamic TS: Parameters}
            \label{tab:DTS_params}
            \centering
            \begin{tabular}{lrrrr}
                \toprule
                                  & Period &    DTS    & dTS           & dOTS       \\
                Environment       & $T$    &   $C$     & $\gamma$      & $\gamma$   \\
                \midrule
                Fast Varying      &   100  &      25   & 0.40     & 0.40    \\
                Slow Varying      &  1000  &     250   & 0.75     & 0.75    \\
                Abruptly Varying  &   250  &      25   & 0.60     & 0.60    \\
                \bottomrule
            \end{tabular}
        \end{table}
        Results are provided in Section 5.1 of main script.

    \subsection*{Discounted-UCB}
        Discounted-UCB is proposed in \cite{Garivier2011b} for non-stationary bandit problems. Specifically, the bandit
        assumed in this case has reward distributions remaining constant over an epoch and which changes at unknown time
        instants. The algorithm works on the principles of upper confidence bound based policies introduced in 
        \cite{Auer2002} and uses a discounting factor $\gamma \in [0,1]$ to reduce the effect of past rewards on 
        current action selection.
        
\begin{figure}[H]
            \begin{subfigure}[b]{0.5\textwidth}
                \includegraphics[width=\linewidth]{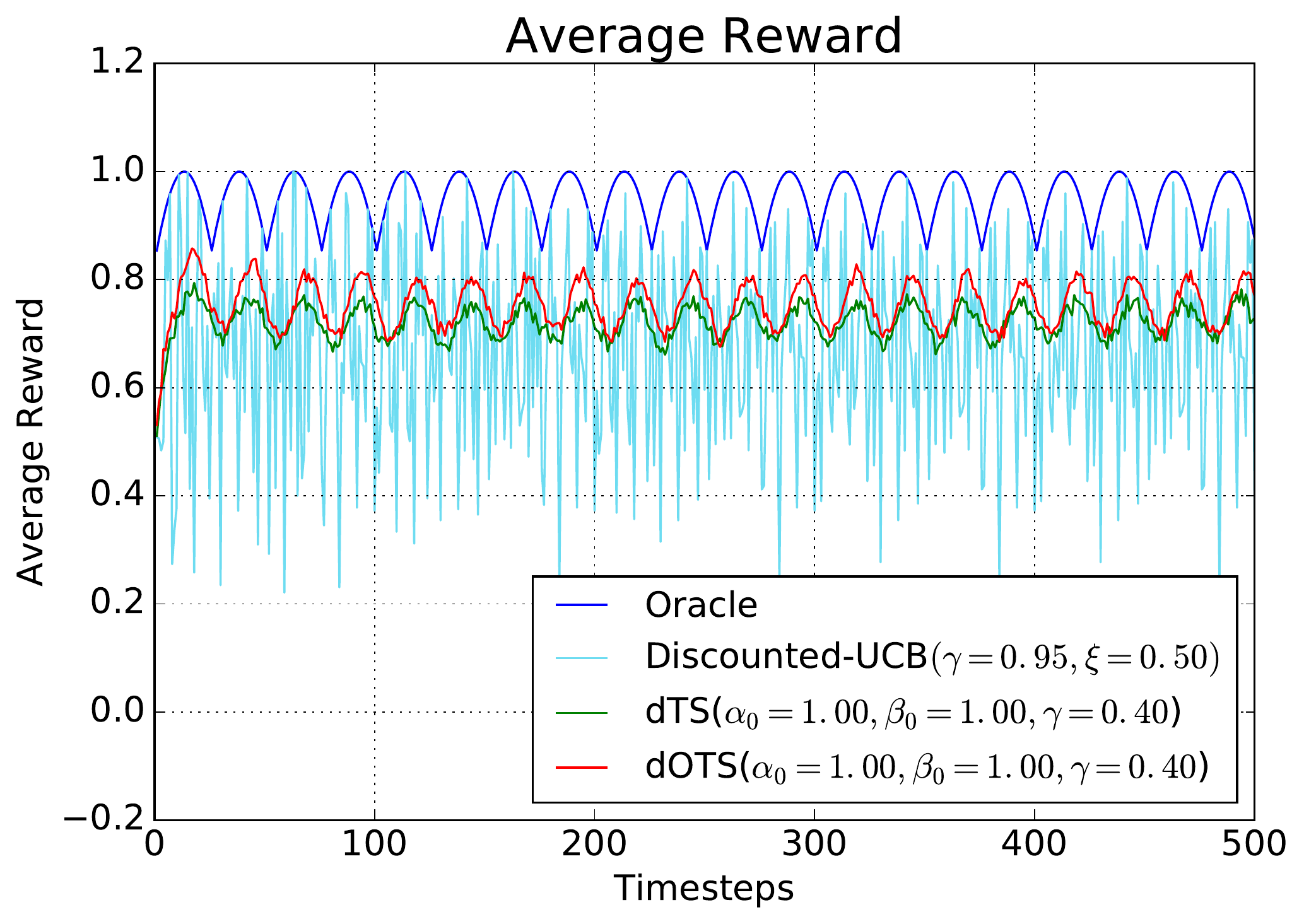}
                \caption{Instantaneous Average Reward}
                \label{fig:DUCB_fve_rewards}
            \end{subfigure}
            \begin{subfigure}[b]{0.5\textwidth}
                \includegraphics[width=\linewidth]{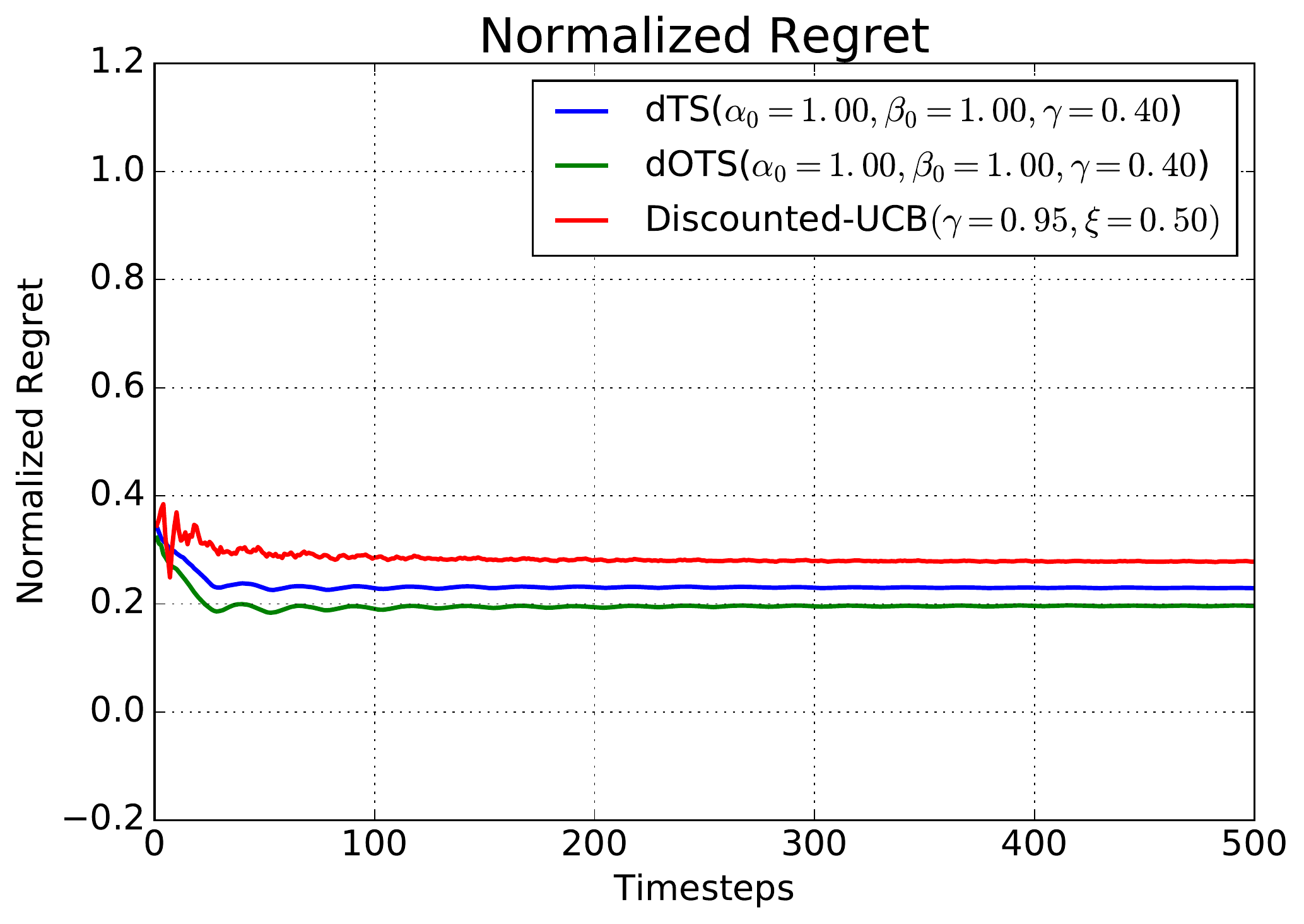}
                \caption{Normalized Regret}
                \label{fig:DUCB_fve_regret}
            \end{subfigure}
            \caption{Comparison against Discounted-UCB in fast varying environment}
            \label{fig:DUCB_fve}
        \end{figure}

\begin{figure}[H]
            \begin{subfigure}[b]{0.5\textwidth}
                \includegraphics[width=\linewidth]{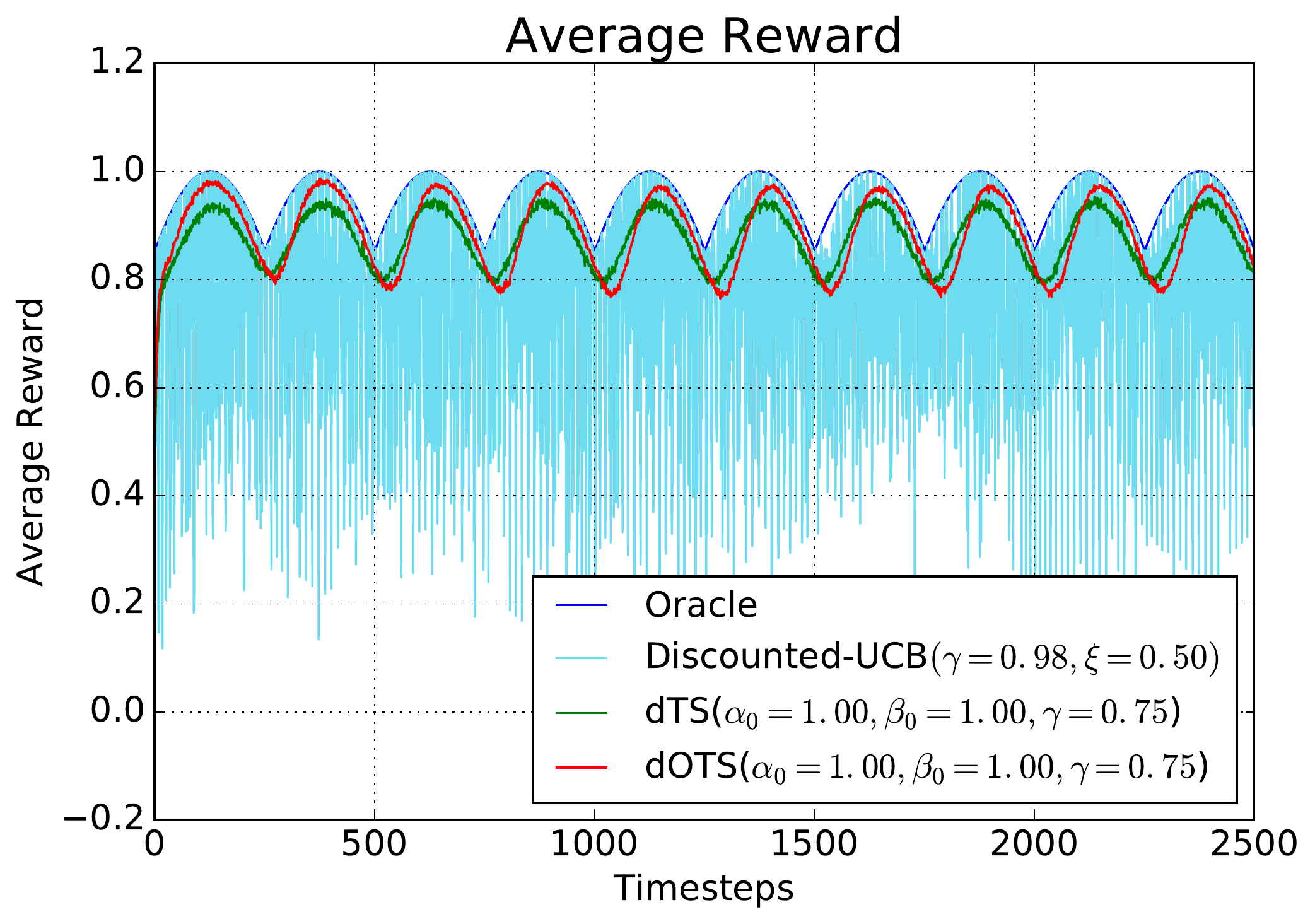}
                \caption{Instantaneous Average Reward}
                \label{fig:DUCB_sve_rewards}
            \end{subfigure}
            \begin{subfigure}[b]{0.5\textwidth}
                \includegraphics[width=\linewidth]{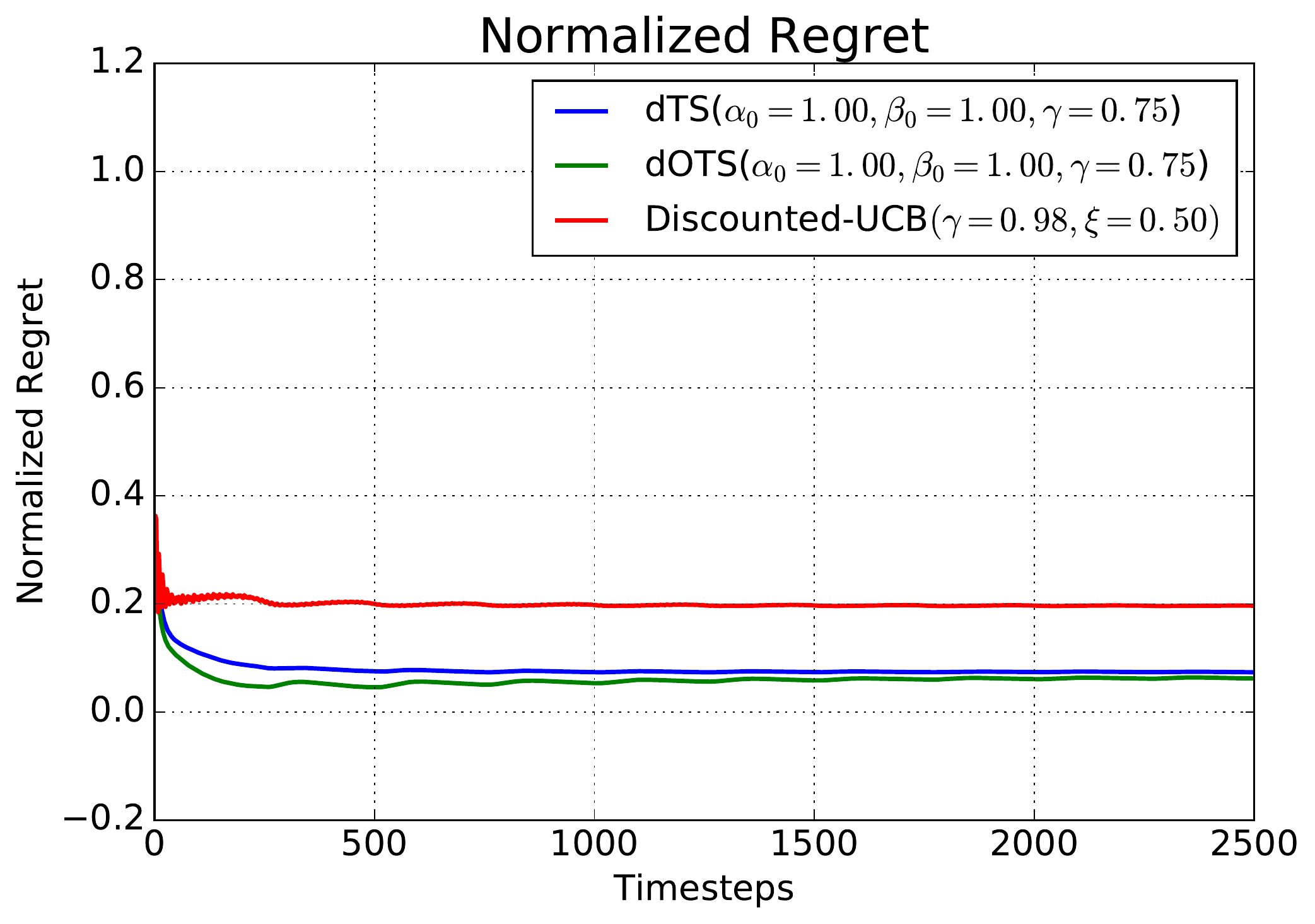}
                \caption{Normalized Regret}
                \label{fig:DUCB_sve_regret}
            \end{subfigure}
            \caption{Comparison against Discounted-UCB in slow varying environment}
            \label{fig:DUCB_sve}
        \end{figure}

\begin{figure}[H]
            \begin{subfigure}[b]{0.5\textwidth}
                \includegraphics[width=\linewidth]{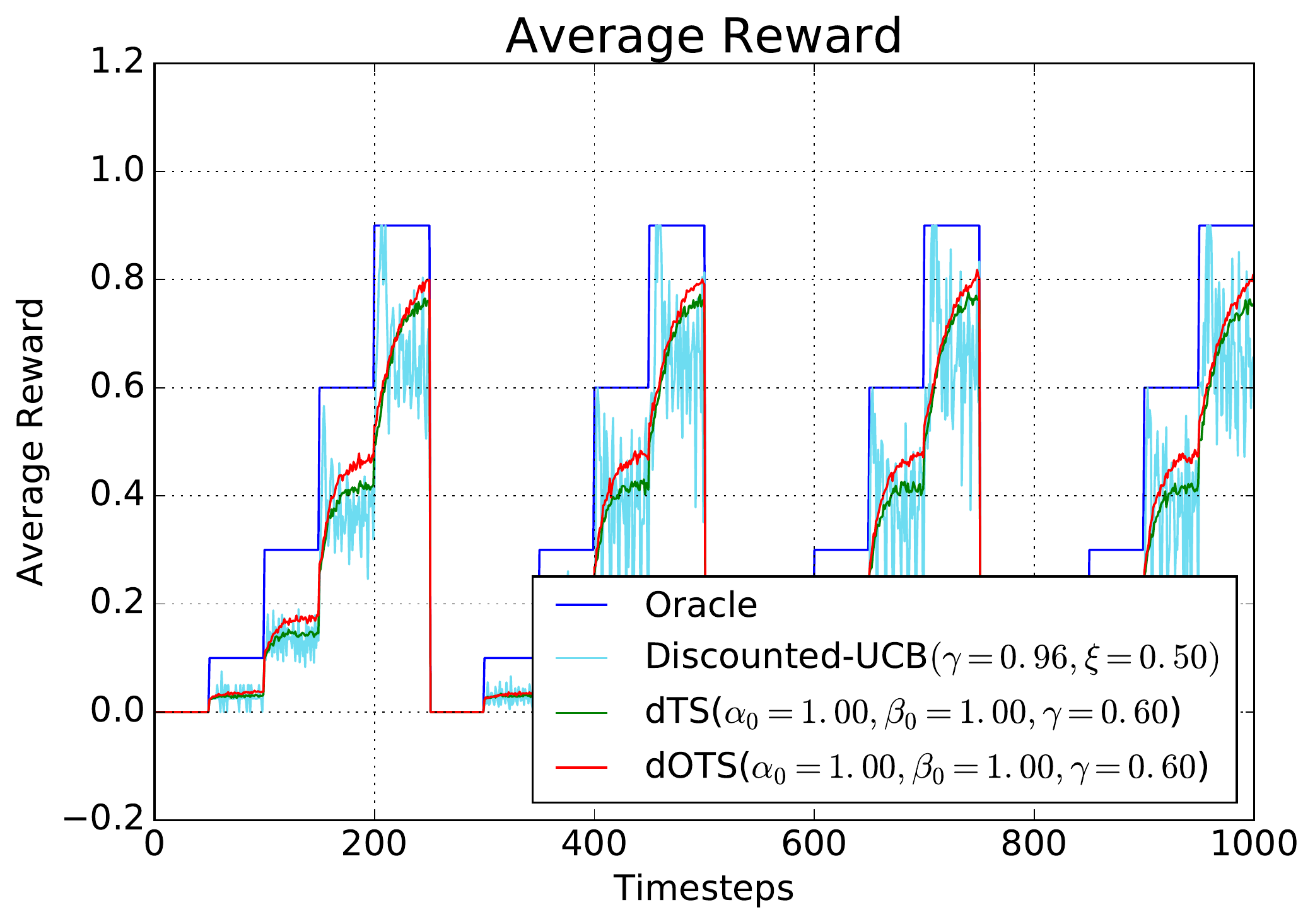}
                \caption{Instantaneous Average Reward}
                \label{fig:DUCB_ave_rewards}
            \end{subfigure}
            \begin{subfigure}[b]{0.5\textwidth}
                \includegraphics[width=\linewidth]{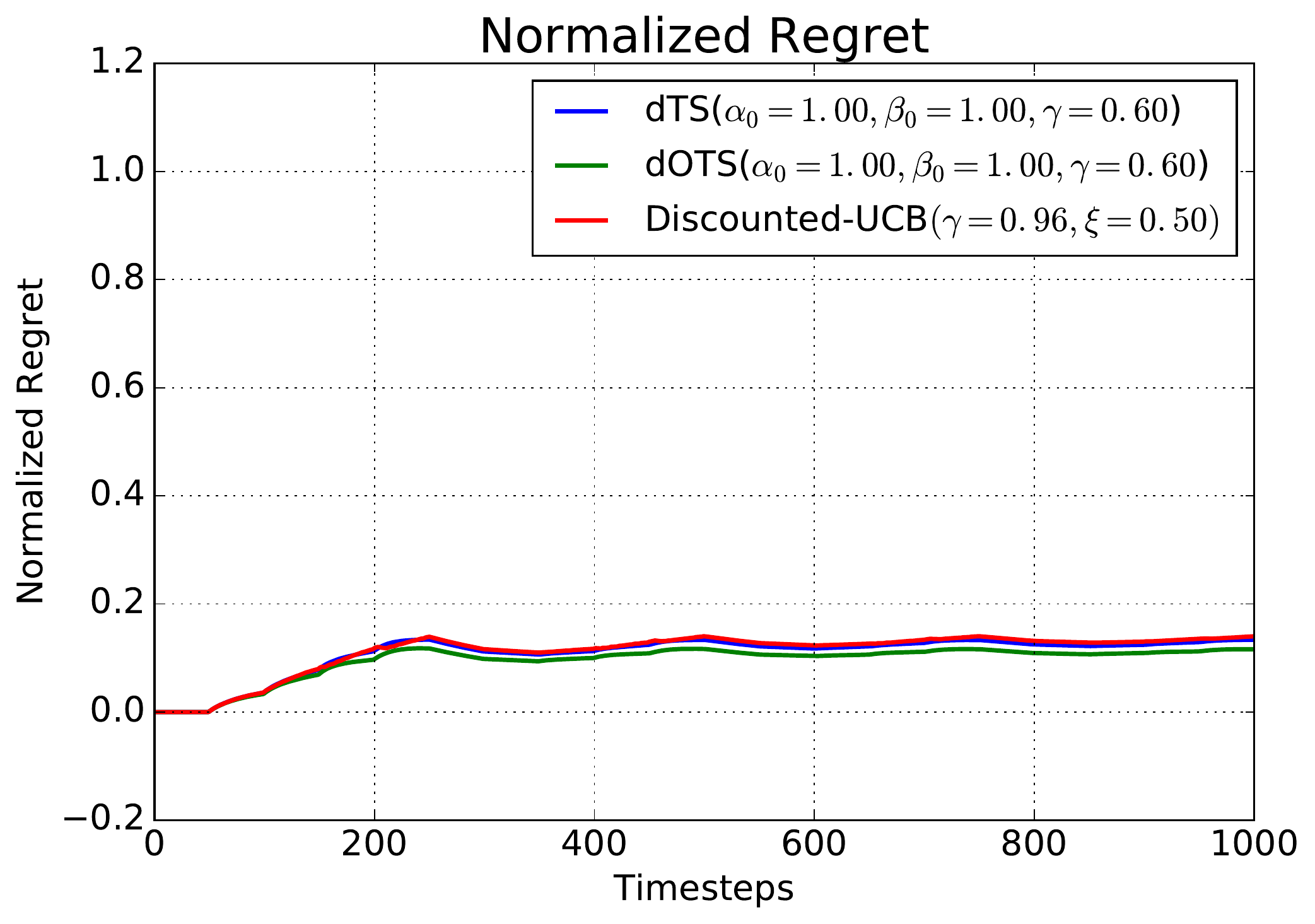}
                \caption{Normalized Regret}
                \label{fig:DUCB_ave_regret}
            \end{subfigure}
            \caption{Comparison against Discounted-UCB in abruptly varying environment}
            \label{fig:DUCB_ave}
        \end{figure}

        For simulation purposes, the discounting factor $\gamma$ is selected according to (\ref{eqn:ducb_gamma}).
        \begin{align}    \label{eqn:ducb_gamma}
            \gamma &= 1 - (4B)^{-1} \sqrt{\frac{\Upsilon_T}{T}}
        \end{align}
        where $\Upsilon_T$ is the number of change points in time time horizon $T$ and $B$ is the bound on the
        reward. Exact values used for simulation is given in Table \ref{tab:ducb_params}. All experiments are conducted
        with $\xi = 0.5$ for D-UCB. Actual values used in simulation are provided in Table \ref{tab:ducb_params} and
        results are shown in Figures \ref{fig:DUCB_fve} - \ref{fig:DUCB_ave}.
        
        \begin{table}[H]
            \caption{Comparison with D-UCB: Parameters (With $B = 1$)}
            \label{tab:ducb_params}
            \centering
            \begin{tabular}{lrrrrr}
                \toprule
                                  & Time Horizon & \multicolumn{2}{c}{D-UCB} & dTS      & dOTS       \\
                Environment       & $T$          &$\Upsilon_T$&  $\gamma$     & $\gamma$ & $\gamma$   \\
                \midrule
                Fast Varying      &   500        &      20    &  0.9500       & 0.40     & 0.40    \\
                Slow Varying      &  2500        &      10    &  0.9842       & 0.75     & 0.75    \\
                Abruptly Varying  &  1000        &      20    &  0.9646       & 0.60     & 0.60    \\
                \bottomrule
            \end{tabular}
        \end{table}

    \subsection*{Sliding Window - UCB}
        Sliding Window UCB (SW-UCB) is also proposed in \cite{Garivier2011b} for non-stationary bandit problems.
        SW-UCB considers only the reward obtained in a past window of time. The length of this window is denoted
        by $\tau$ and is calculated as
    
        \begin{align}
            \tau &= 2B \sqrt{\frac{T log T}{\Upsilon_T}}.
        \end{align}
        
\begin{figure}[H]
            \begin{subfigure}[b]{0.5\textwidth}
                \includegraphics[width=\linewidth]{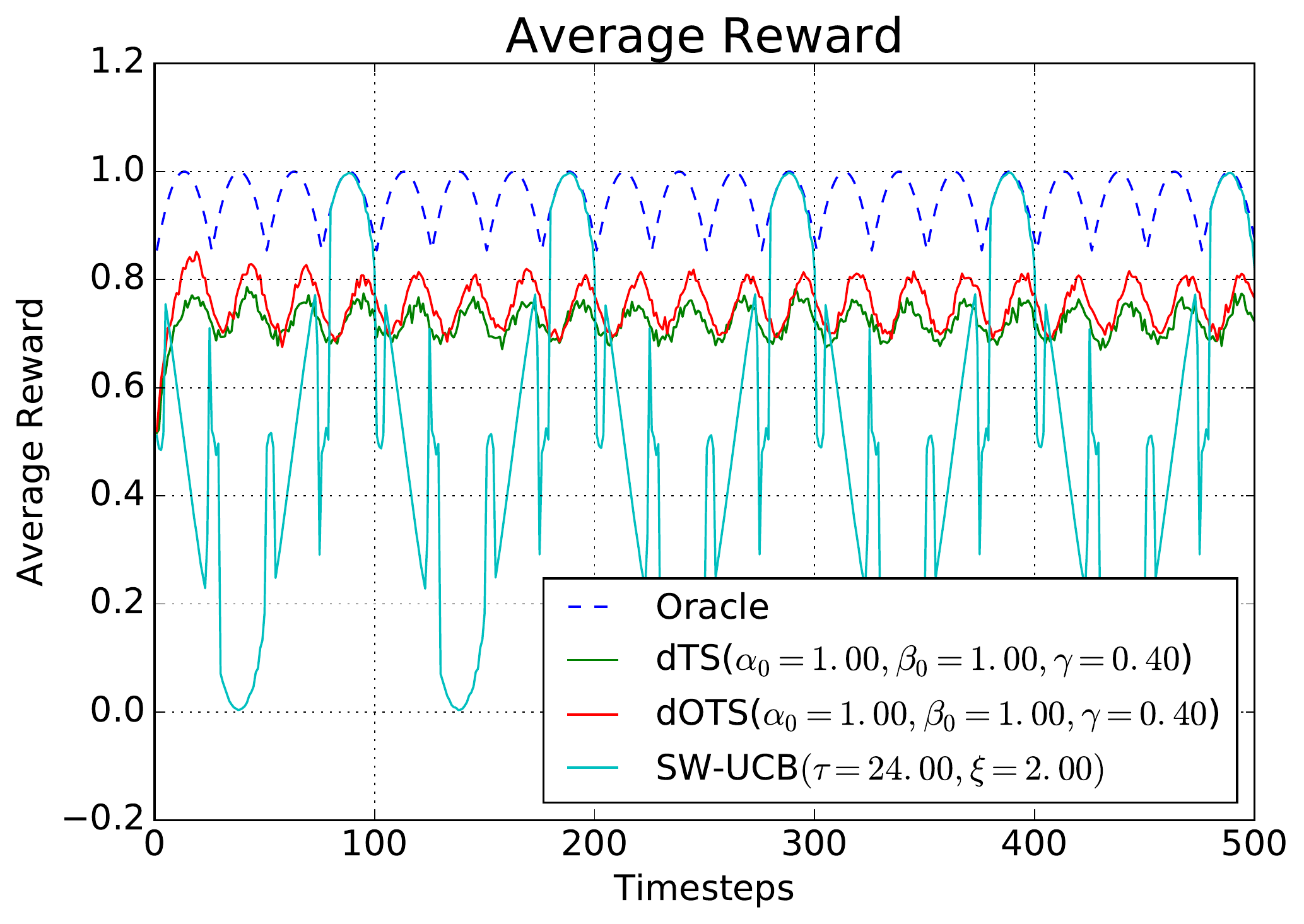}
                \caption{Instantaneous Average Reward}
                \label{fig:SWUCB_fve_rewards}
            \end{subfigure}
            \begin{subfigure}[b]{0.5\textwidth}
                \includegraphics[width=\linewidth]{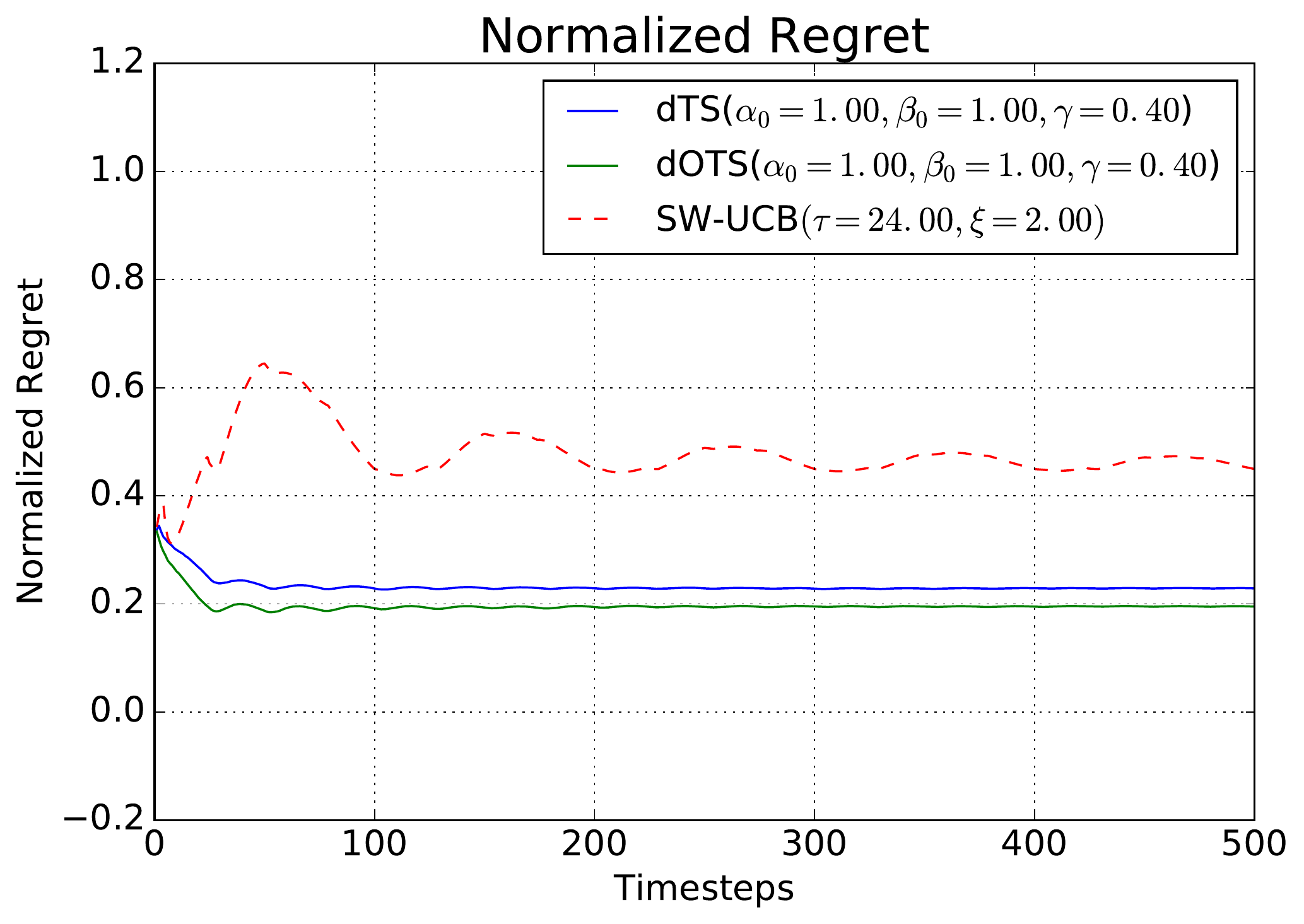}
                \caption{Normalized Regret}
                \label{fig:SWUCB_fve_regret}
            \end{subfigure}
            \caption{Comparison against Sliding Window UCB in fast varying environment}
            \label{fig:SWUCB_fve}
        \end{figure}

\begin{figure}[H]
            \begin{subfigure}[b]{0.5\textwidth}
                \includegraphics[width=\linewidth]{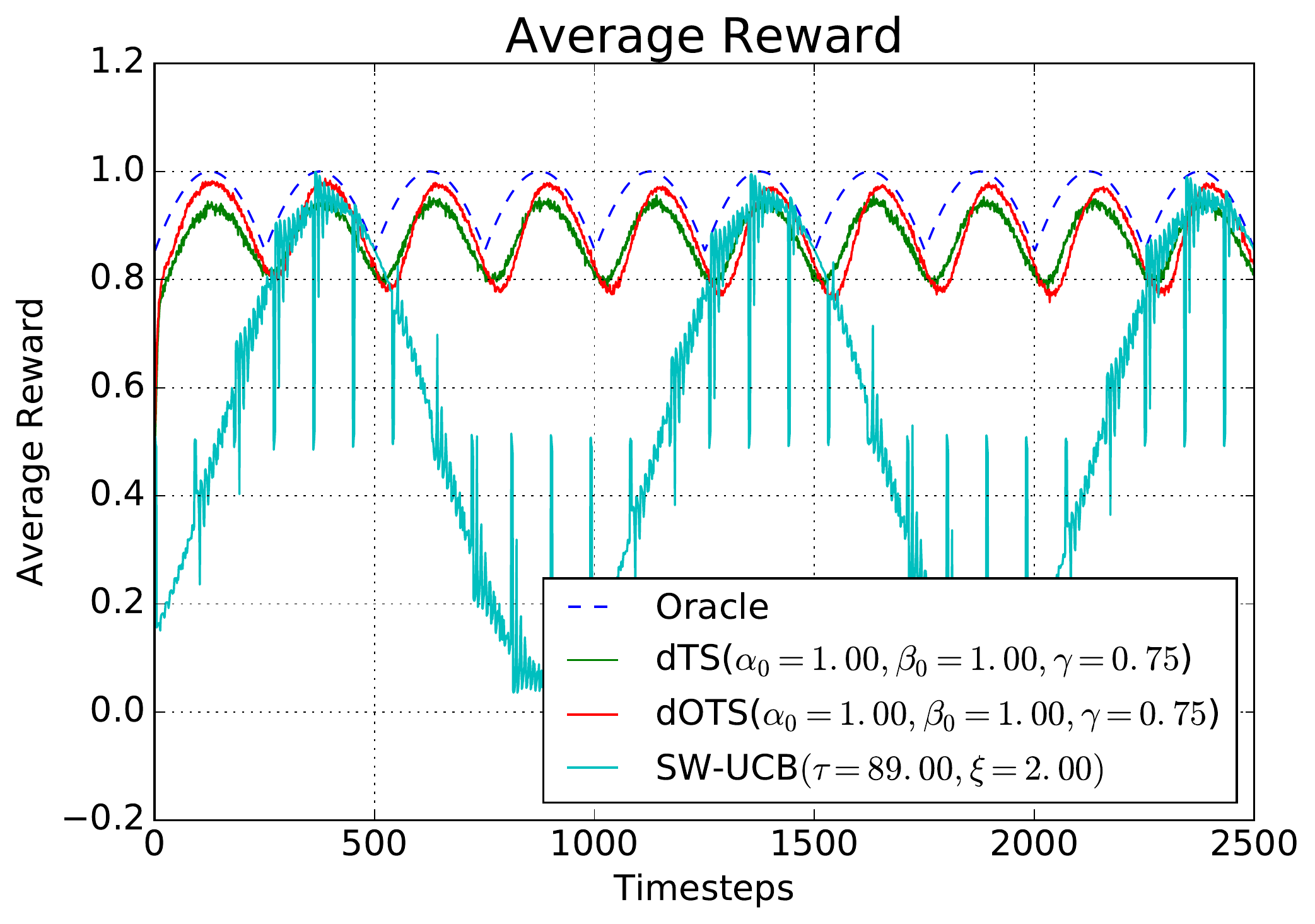}
                \caption{Instantaneous Average Reward}
                \label{fig:SWUCB_sve_rewards}
            \end{subfigure}
            \begin{subfigure}[b]{0.5\textwidth}
                \includegraphics[width=\linewidth]{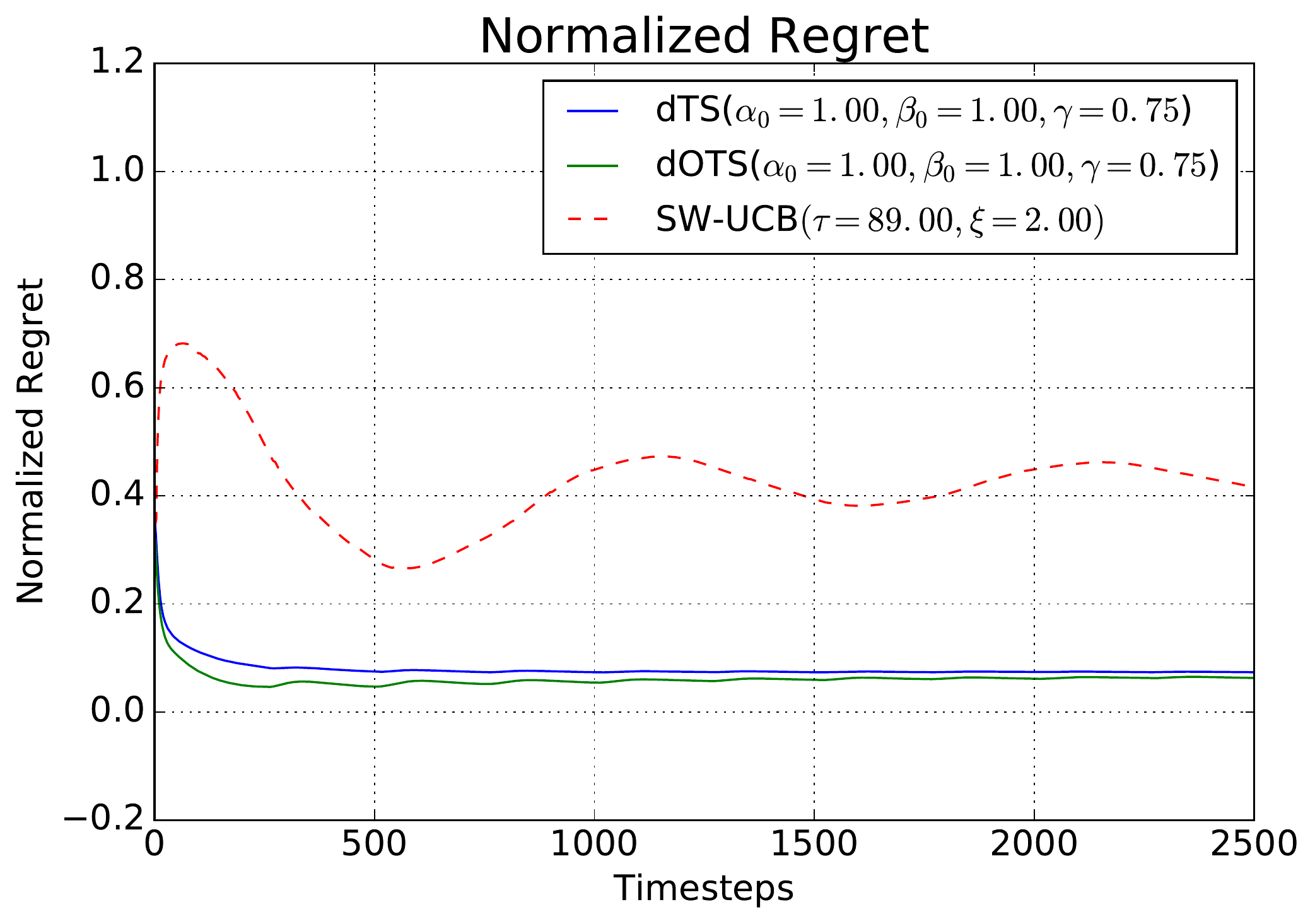}
                \caption{Normalized Regret}
                \label{fig:SWUCB_sve_regret}
            \end{subfigure}
            \caption{Comparison against Sliding Window UCB in slow varying environment}
            \label{fig:SWUCB_sve}
        \end{figure}

\begin{figure}[H]
            \begin{subfigure}[b]{0.5\textwidth}
                \includegraphics[width=\linewidth]{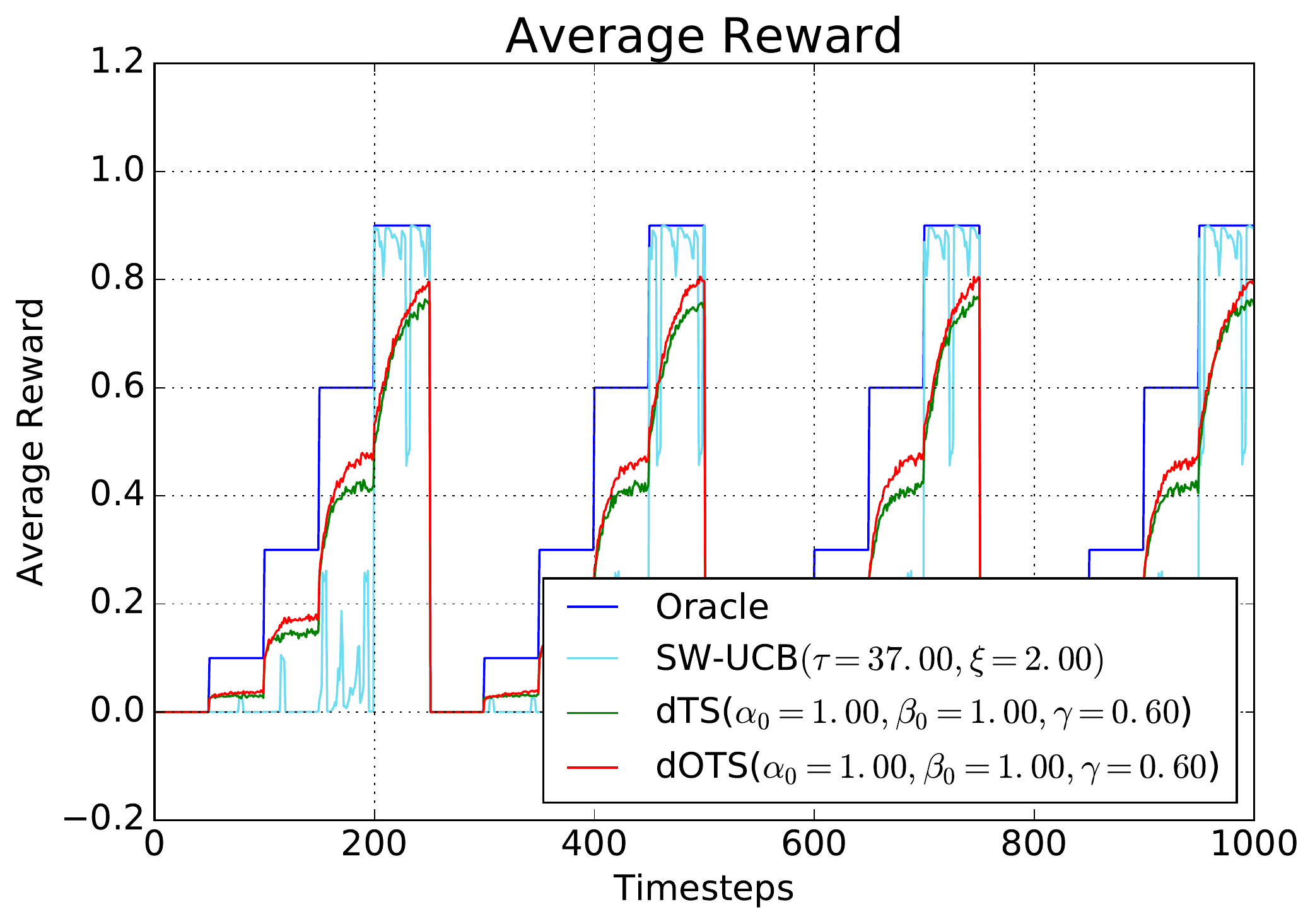}
                \caption{Instantaneous Average Reward}
                \label{fig:SWUCB_ave_rewards}
            \end{subfigure}
            \begin{subfigure}[b]{0.5\textwidth}
                \includegraphics[width=\linewidth]{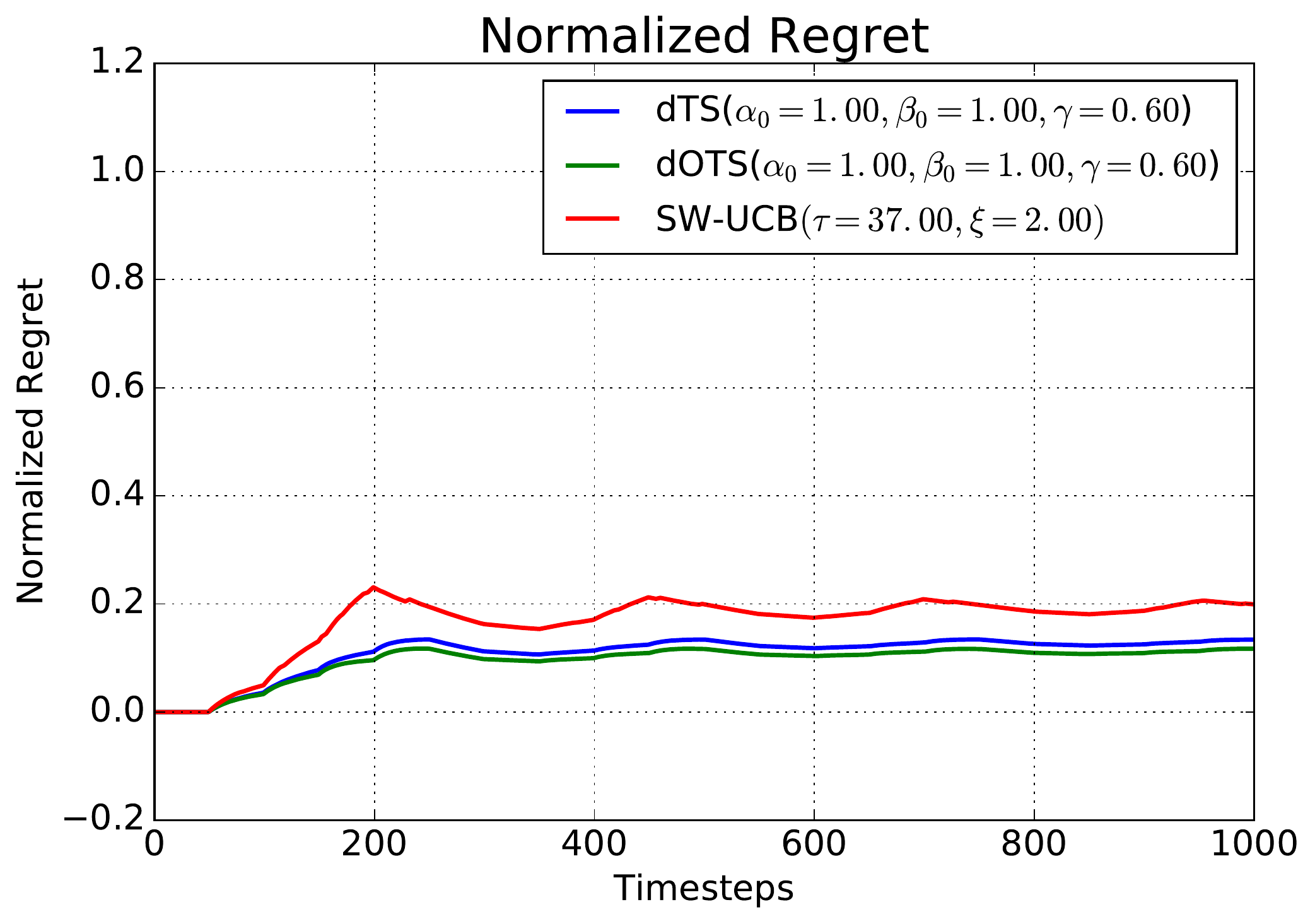}
                \caption{Normalized Regret}
                \label{fig:SWUCB_ave_regret}
            \end{subfigure}
            \caption{Comparison against Sliding Window UCB in abruptly varying environment}
            \label{fig:SWUCB_ave}
        \end{figure}

        Table \ref{tab:swucb_params} contains the actual values used for simulation and results are shown in
        Figures \ref{fig:SWUCB_fve}-\ref{fig:SWUCB_ave}.
        
        \begin{table}[H]
            \caption{Comparison with SW-UCB: Parameters (With $B = 1$)}
            \label{tab:swucb_params}
            \centering
            \begin{tabular}{lrrrrr}
                \toprule
                                  & Time Horizon & \multicolumn{2}{c}{SW-UCB} & dTS      & dOTS       \\
                Environment       & $T$          &$\Upsilon_T$&  $\tau$     & $\gamma$ & $\gamma$   \\
                \midrule
                Fast Varying      &   500        &      20    &  24       & 0.40     & 0.40    \\
                Slow Varying      &  2500        &      10    &  89       & 0.75     & 0.75    \\
                Abruptly Varying  &  1000        &      20    &  37       & 0.60     & 0.60    \\
                \bottomrule
            \end{tabular}
        \end{table}
        
    \subsection*{EXP3-IX}
        EXP3-IX is proposed in \cite{Neu2015a} for non-stochastic bandits. It showed that, explicit exploration
        is not necessary to achieve high probability regret bounds in non-stochastic bandits. EXP3-IX has two
        non negative parameters - $\eta$, similar to the egalitarianism parameter in EXP3 and $\gamma$, which
        is the implicit exploration parameter. For high probability performance bounds, the values of 
        $\eta$ and $\gamma$ is calculated as
        \begin{align}
            \eta    = \sqrt{\frac{2 log K}{K T}} \qquad \text{and} \qquad
            \gamma  = \frac{\eta}{2}
        \end{align}
        
\begin{figure}[H]
            \begin{subfigure}[b]{0.5\textwidth}
                \includegraphics[width=\linewidth]{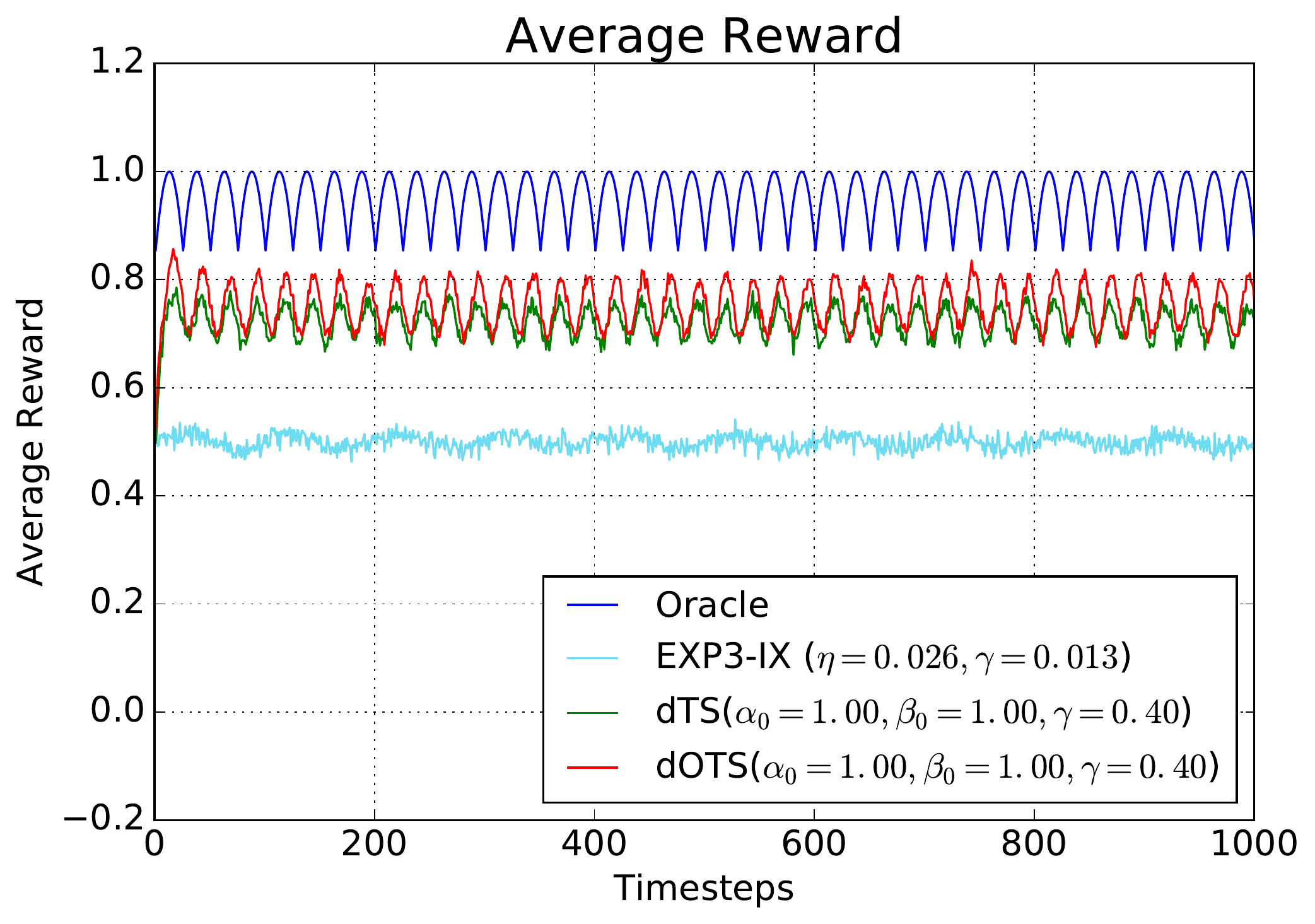}
                \caption{Instantaneous Average Reward}
                \label{fig:EXP3IX_fve_rewards}
            \end{subfigure}
            \begin{subfigure}[b]{0.5\textwidth}
                \includegraphics[width=\linewidth]{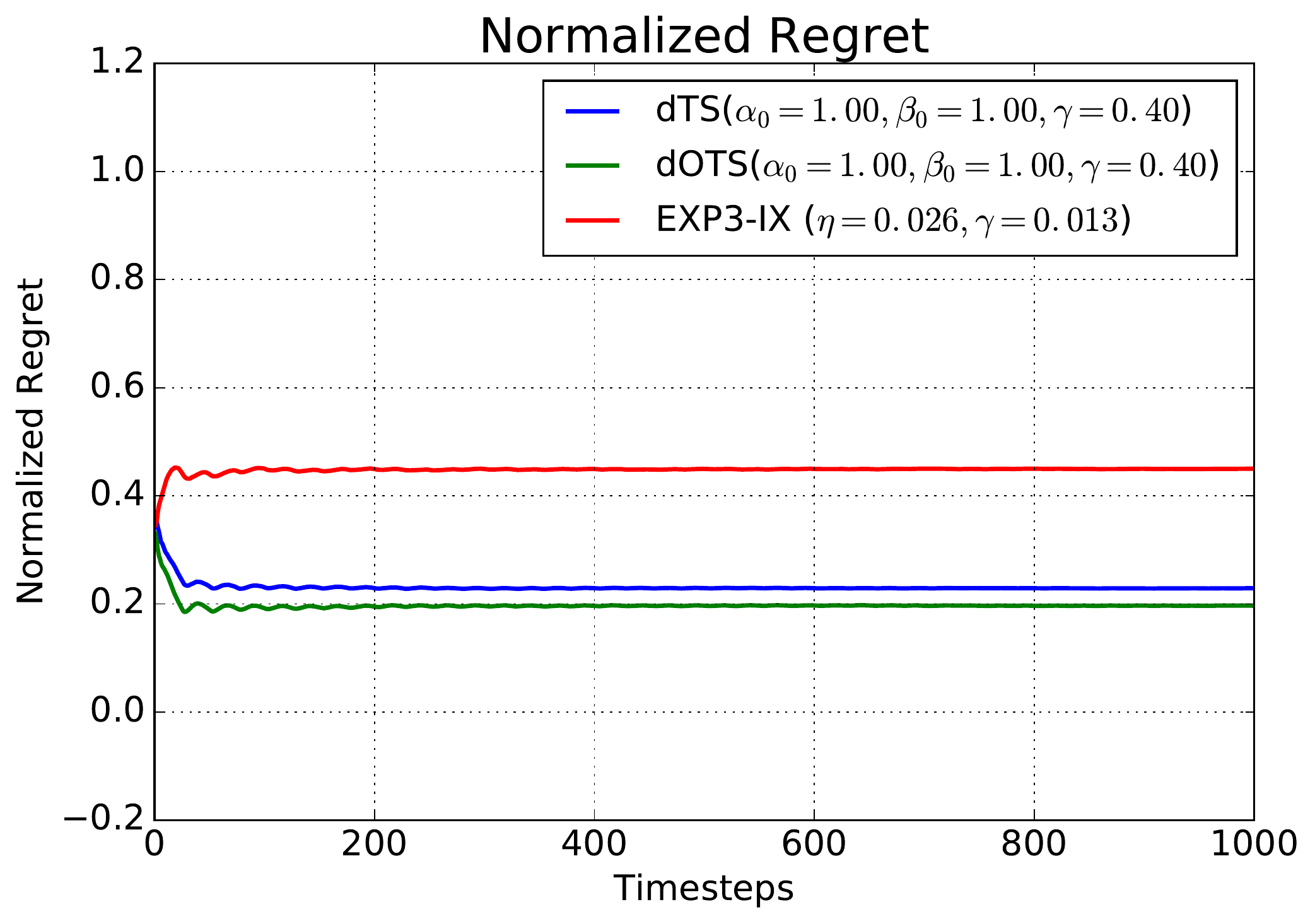}
                \caption{Normalized Regret}
                \label{fig:EXP3IX_fve_regret}
            \end{subfigure}
            \caption{Comparison against EXP3-IX in fast varying environment}
            \label{fig:EXP3IX_fve}
        \end{figure}

\begin{figure}[H]
            \begin{subfigure}[b]{0.5\textwidth}
                \includegraphics[width=\linewidth]{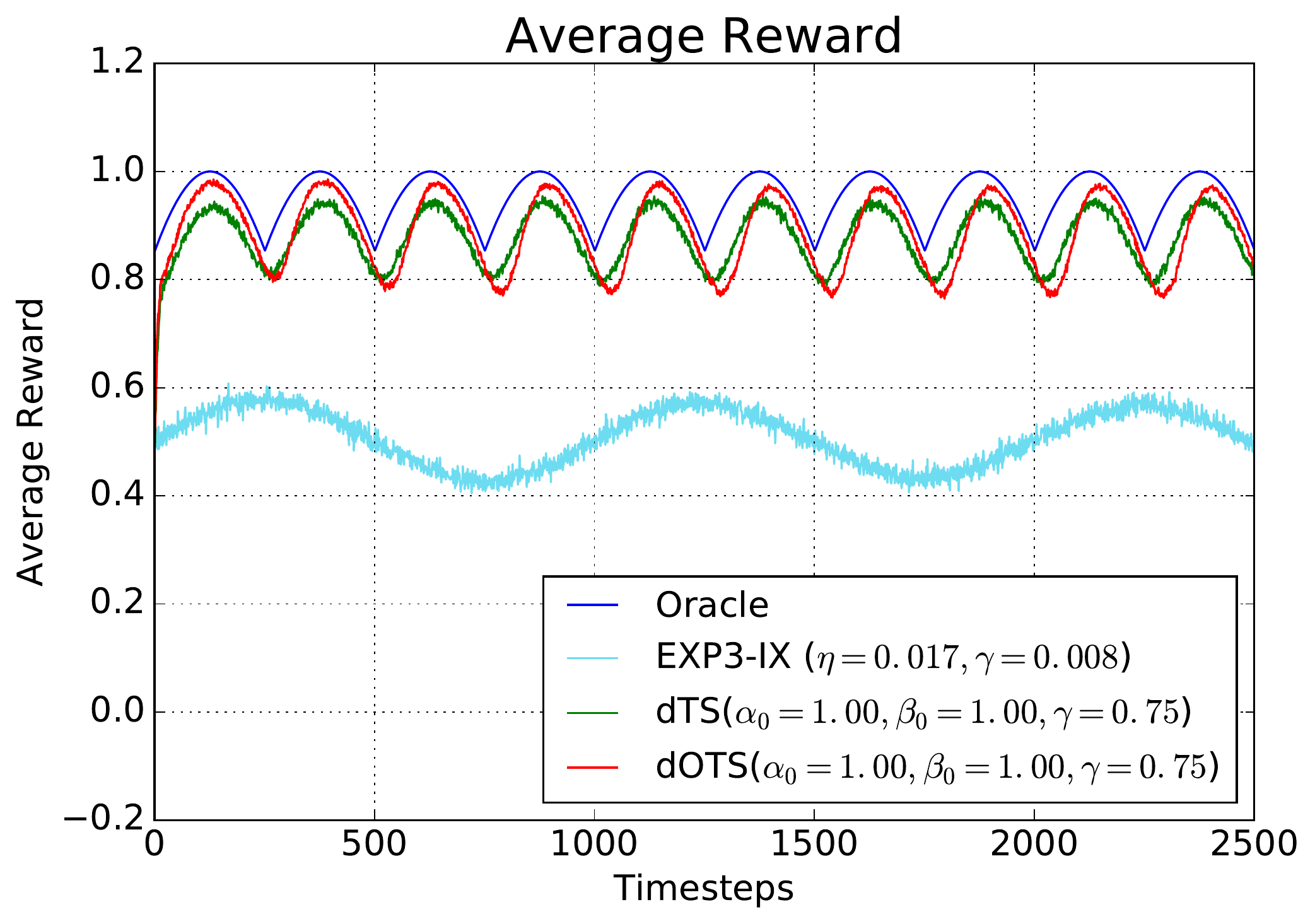}
                \caption{Instantaneous Average Reward}
                \label{fig:EXP3IX_sve_rewards}
            \end{subfigure}
            \begin{subfigure}[b]{0.5\textwidth}
                \includegraphics[width=\linewidth]{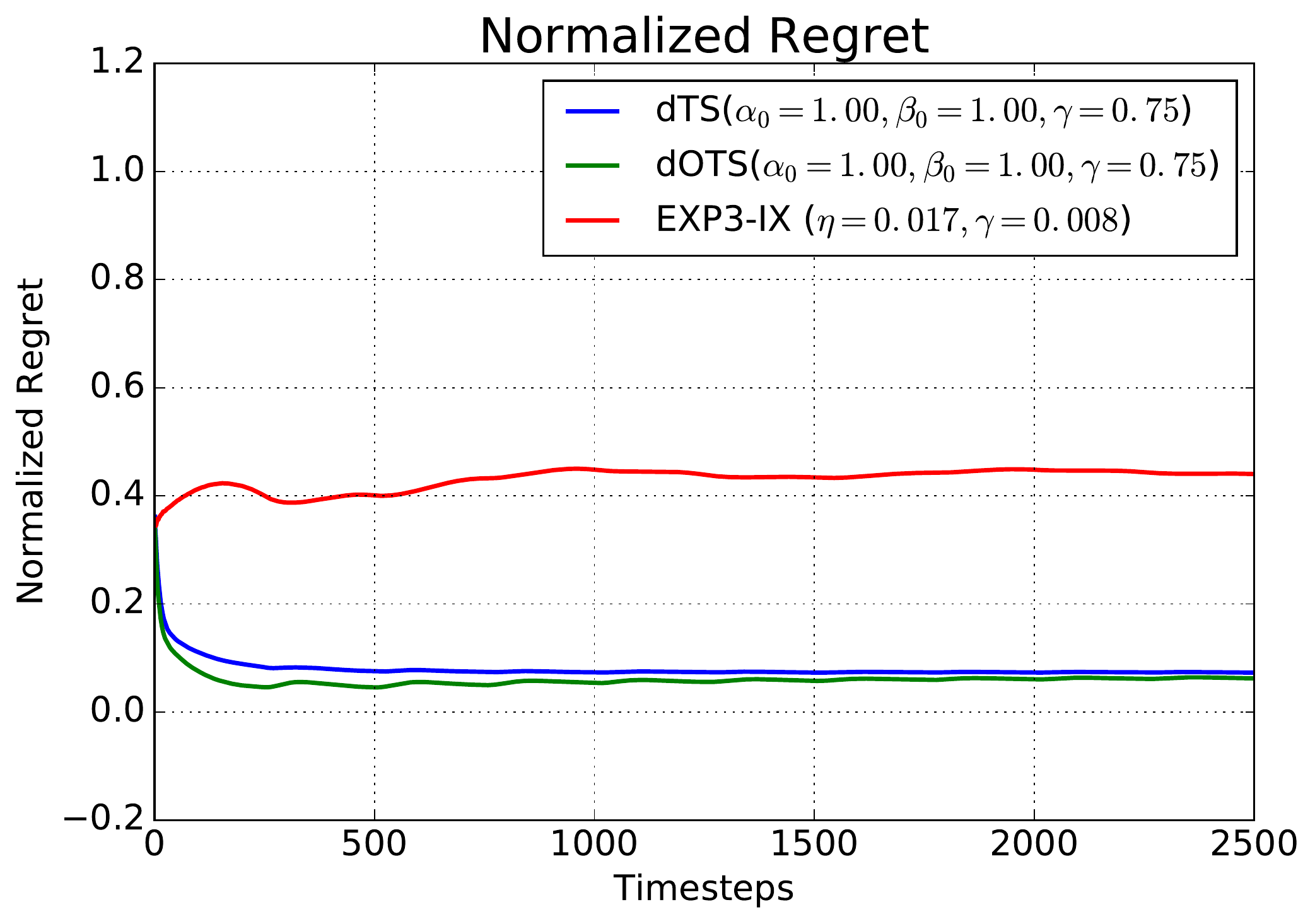}
                \caption{Normalized Regret}
                \label{fig:EXP3IX_sve_regret}
            \end{subfigure}
            \caption{Comparison against EXP3-IX in slow varying environment}
            \label{fig:EXP3-IX_sve}
        \end{figure}

\begin{figure}[H]
            \begin{subfigure}[b]{0.5\textwidth}
                \includegraphics[width=\linewidth]{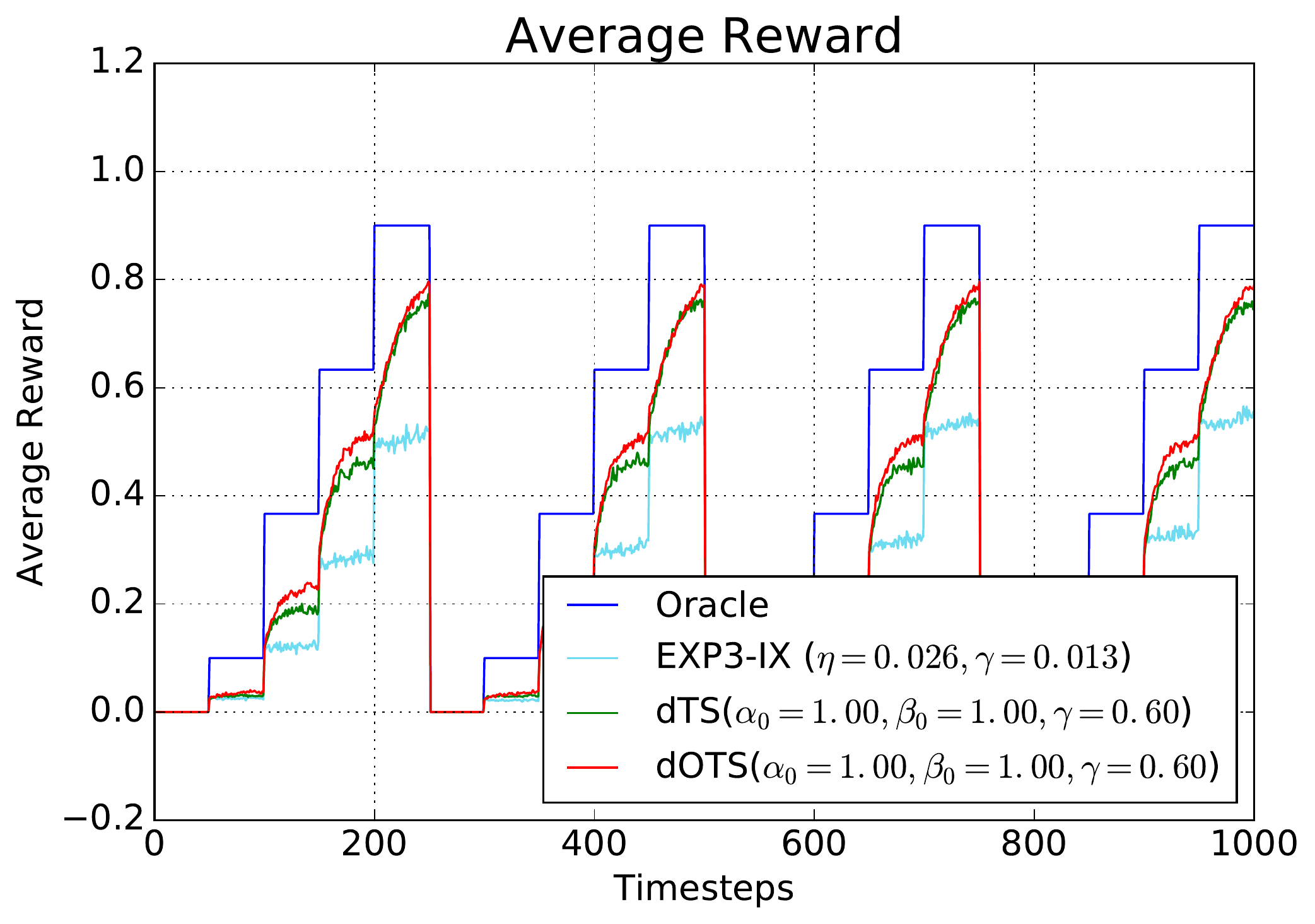}
                \caption{Instantaneous Average Reward}
                \label{fig:EXP3IX_ave_rewards}
            \end{subfigure}
            \begin{subfigure}[b]{0.5\textwidth}
                \includegraphics[width=\linewidth]{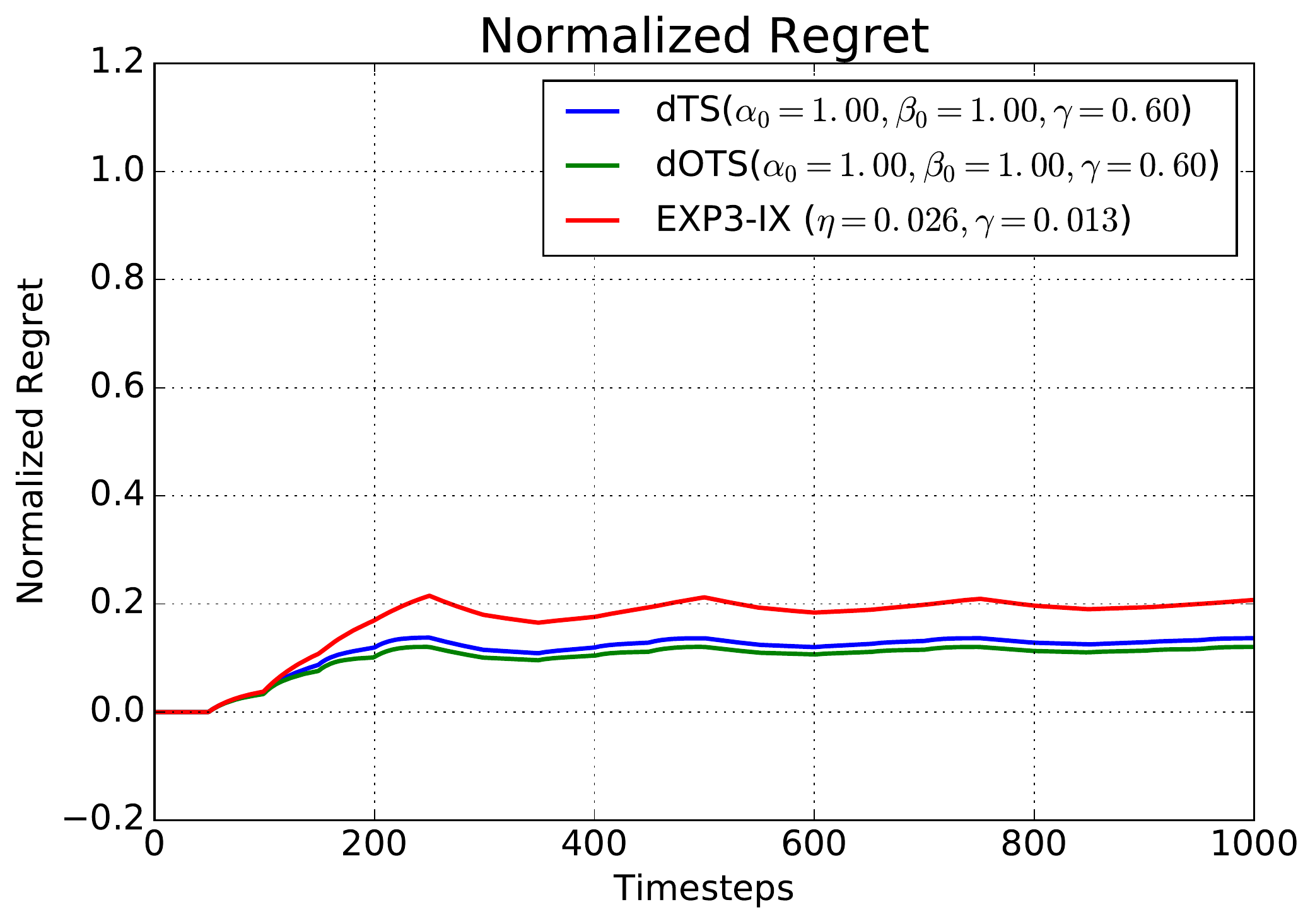}
                \caption{Normalized Regret}
                \label{fig:EXP3IX_ave_regret}
            \end{subfigure}
            \caption{Comparison against EXP3-IX in abruptly varying environment}
            \label{fig:EXP3IX_ave}
        \end{figure}

        Table \ref{tab:swucb_params} contains the actual values used for simulation and results are shown in
        Figures \ref{fig:EXP3IX_fve}-\ref{fig:EXP3IX_ave}.
        
        \begin{table}[H]
            \caption{Comparison with SW-UCB: Parameters (With $B = 1$)}
            \label{tab:exp3ix_params}
            \centering
            \begin{tabular}{lrrrrr}
                \toprule
                                  & Time Horizon & \multicolumn{2}{c}{EXP3-IX} & dTS      & dOTS       \\
                Environment       & $T$          &    $\eta$  &  $\gamma$ & $\gamma$ & $\gamma$   \\
                \midrule
                Fast Varying      &   500        &    0.0263  &  0.0132   & 0.40     & 0.40    \\
                Slow Varying      &  2500        &    0.01665 &  0.00832  & 0.75     & 0.75    \\
                Abruptly Varying  &  1000        &    0.0263  &  0.0132   & 0.60     & 0.60    \\
                \bottomrule
            \end{tabular}
        \end{table}
        
    \subsection*{Discussion}
        From all results shown above, we can observe that dTS and dOTS is able to perform better than
        the state of the art algorithms. One interesting case in these experiments is the behaviour of
        Discounted-UCB in abruptly varying environment (Figure \ref{fig:DUCB_ave}). We can see that
        the regret of DUCB is almost comparable to that of dTS, still dOTS being the better. DUCB also
        uses similar type of discounting as used in dTS and that could be the reason for its good
        performance in abruptly varying environment. But in case of slow varying and fast varying
        environments, this effect is not observed.
        
        Even though EXP3-IX is proposed for a non-stochastic environment, its regret is the highest
        among all the algorithms compared, for fast and slow varying cases. It's worth exploring
        the impact of Implicit Exploration factor ($\gamma$) in this context to analyse the cause.

\end{document}